\documentclass{article}

\usepackage[nonatbib,final]{nips_2018}

\usepackage[utf8]{inputenc} 
\usepackage[T1]{fontenc}    
\usepackage{hyperref}       
\usepackage{url}            
\usepackage{cite}
\usepackage{booktabs}       
\usepackage{amsmath}
\usepackage{amsfonts}       
\usepackage{nicefrac}       
\usepackage{microtype}      
\usepackage{algorithm}
\usepackage{algorithmic}
\usepackage{xspace}

\usepackage{epsfig}
\usepackage{graphicx}
\usepackage{amsmath}
\usepackage{amssymb}
\usepackage{caption}
\usepackage{subcaption}
\usepackage{pifont}
\usepackage{xcolor}
\usepackage[compact]{titlesec}
\usepackage{epstopdf}
\usepackage{times}
\usepackage{latexsym}
\usepackage{enumitem}
\usepackage{multirow}
\usepackage{array}
\usepackage{dsfont}
\usepackage{boldline}
\usepackage{dblfloatfix}

\newcommand{\calE}{\mathcal{E}}

\newcommand{\calD}{\mathcal{D}}

\newcommand{\calL}{\mathcal{L}}
\newcommand{\calA}{\mathcal{A}}

\newcommand{\calT}{\mathcal{T}}

\newcommand{\gridworld}{\textsc{gridworld}\xspace}
\newcommand{\thor}{\textsc{thor}\xspace}

\newcommand{\etal}{\textit{et al}.\xspace}
\newcommand{\ie}{\textit{i}.\textit{e}.\xspace}
\newcommand{\eg}{\textit{e}.\textit{g}.\xspace}

\newcommand{\env}{$\varepsilon$}
\newcommand{\task}{$\tau$}

\newcommand{\ourmethod}{$\textsc{SynPo}$\xspace}
\newcommand{\rollout}{\textsc{rollout}\xspace}

\usepackage{amsmath,amsfonts}
\usepackage{amssymb,amsopn}
\usepackage{bm} 

\usepackage{amssymb}
\usepackage{amsmath,amsfonts}
\usepackage{amsthm,amsopn}

\usepackage{bm} 

\newcommand{\vct}[1]{\boldsymbol{#1}} 
\newcommand{\mat}[1]{\boldsymbol{#1}} 

\newcommand{\field}[1]{\mathbb{#1}}
\newcommand{\R}{\field{R}} 
\newcommand{\T}{^{\textrm T}} 

\newcommand{\ProbOpr}[1]{\mathbb{#1}}

\newcommand{\expect}[2]{%
\ifthenelse{\equal{#2}{}}{\ProbOpr{E}_{#1}}
{\ifthenelse{\equal{#1}{}}{\ProbOpr{E}\left[#2\right]}{\ProbOpr{E}_{#1}\left[#2\right]}}} 
\newcommand{\var}[2]{%
\ifthenelse{\equal{#2}{}}{\ProbOpr{VAR}_{#1}}
{\ifthenelse{\equal{#1}{}}{\ProbOpr{VAR}\left[#2\right]}{\ProbOpr{VAR}_{#1}\left[#2\right]}}} 

\newcommand{\mTheta}{\vct{\Theta}}

\newcommand{\mU}{\mat{U}}
\newcommand{\mV}{\mat{V}}

\newcommand{\vx}{{\vct{x}}}

\newcommand{\vphi}{\vct{\phi}}
\newcommand{\vpsi}{\vct{\psi}}

\newcommand{\eat}[1]{}

\title{Synthesized Policies for Transfer and Adaptation \\ across Tasks and Environments}

\author{
	\textbf{Hexiang Hu}~\thanks{\scriptsize Equal Contribution.} \\
	University of Southern California \\
	Los Angeles, CA 90089 \\ 
	\texttt{hexiangh@usc.edu} \\
	\and
	\textbf{Liyu Chen}~\footnotemark[1]\\
	University of Southern California \\
	Los Angeles, CA 90089 \\ 
	\texttt{liyuc@usc.edu} \\
	\vspace{1pt}
	\and
	\textbf{Boqing Gong}\\
	Tencent AI Lab \\
	Bellevue, WA 98004 \\
	\texttt{boqinggo@outlook.com} \\
	\and
	\textbf{Fei Sha}~\thanks{\scriptsize On leave from University of Southern California (feisha@usc.edu).}\\
	Netflix \\
	Los Angeles, CA 90028 \\
	\texttt{fsha@netflix.com} \\
}

\begin{document}

\maketitle

\begin{abstract}


The ability to transfer in reinforcement learning is key towards building an agent of general artificial intelligence. In this paper, we consider the problem of learning to simultaneously transfer across both environments (\env) and tasks (\task), probably more importantly, by learning from only sparse (\env, \task) pairs out of all the possible combinations. We propose a novel compositional neural network architecture which depicts a meta rule for composing policies from  environment and task embeddings. Notably, one of the main challenges is to learn the embeddings jointly with the meta rule. We further propose new training methods to disentangle the embeddings, making them both distinctive signatures of the environments and tasks and effective building blocks for composing the policies. Experiments on \gridworld and \thor, of which the agent takes as input an egocentric view, show that our approach gives rise to high success rates on all the (\env, \task) pairs after learning from only 40\% of them.

\end{abstract}

\section{Introduction}
\label{sIntro}

Remarkable progress has been made in reinforcement learning in the last few years~\cite{Silver2017MasteringCA,Mnih2015HumanlevelCT,vinyals2017starcraft}.
Among these, an agent learns to discover its best policy of actions to accomplish a task, by interacting with the environment. However, the skills the agent learns are often tied for a specific pair of the environment (\env) and the task (\task). Consequently, when the environment changes even slightly, the agent's performance deteriorates drastically~\cite{overfit,huang2017adversarial}. Thus, being able to swiftly adapt to new environments and transfer skills to new tasks is crucial for the agents to act in real-world settings. 

\emph{How can we achieve swift adaptation and transfer?} In this paper, we consider several progressively difficult settings. In the first setting, the agent needs to adapt and transfer to a new pair of environment and task, when the agent has been exposed to the environment and the task before (but not simultaneously). Our goal is to use as few as possible \emph{seen} pairs (\ie, a subset out of all possible (\env, \task) combinations, as sparse as possible) to train the agent. 

In the second setting, the agent needs to adapt and transfer  across either environments {or} tasks, to those previously unseen by the agent.
For instance, a home service robot needs to adapt from one home to another one but essentially accomplish the same sets of tasks, or the robot learns new tasks in the same home. In the third setting, the agent has encountered neither the environment nor the task before. Intuitively, the second and the third settings are much more challenging than the first one and appear to be  intractable. Thus, the agent is allowed to have a very limited amount of learning data in the target environment and/or task, for instance, from one demonstration, in order to transfer knowledge from its prior learning.

\begin{figure}[t]
	\centering
	\includegraphics[width=0.995\textwidth]{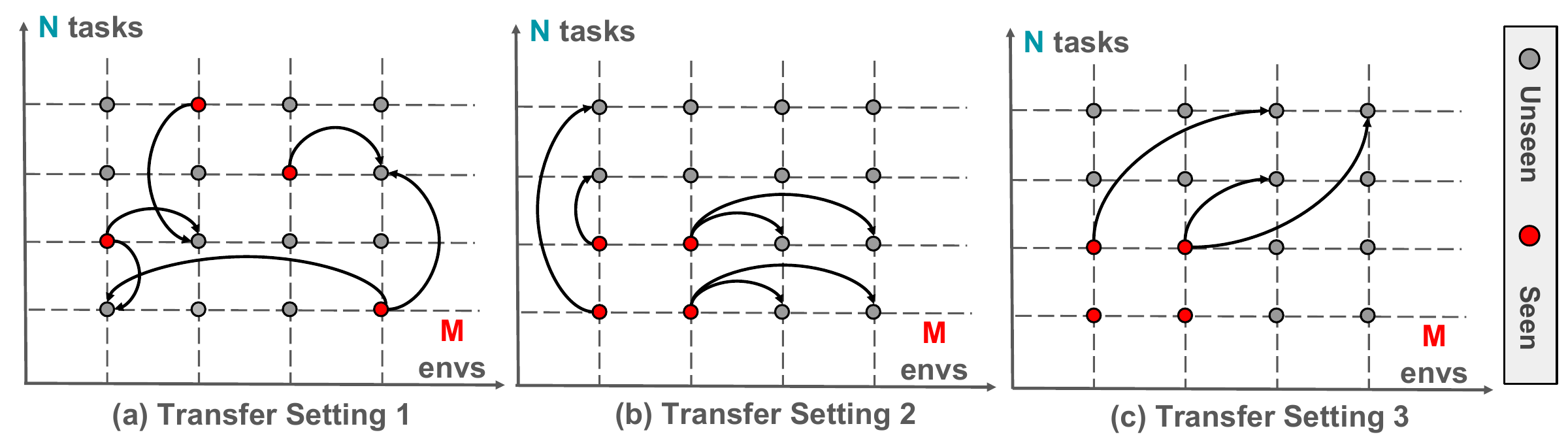}
	\caption{ \small  We consider a transfer learning scenario in reinforcement learning that considers transfer in both task and environment. Three different settings are presented here (see text for details). The \textcolor{red}{\textbf{red dots}} denote \textsc{seen} combinations, \textcolor{gray}{\textbf{gray dots}} denote \textsc{unseen} combinations, and arrows $\rightarrow$ denote transfer directions.}
	\label{fig:illustration}
\end{figure}

Figure~\ref{fig:illustration} schematically illustrates the three settings. Several existing approaches have been proposed to address some of those settings~\cite{distral,Taylor2009TransferLF,oh2017zero,Andreas2017ModularMR,module,Kulkarni2016DeepSR,Barreto2017SuccessorFF}; for a detailed discussion, see related works in Section~\ref{sRelated}.  A common strategy behind these works is to jointly learn through multi-task (reinforcement) learning~\cite{actormimic, distill,distral}. Despite many progresses, however, adaptation and transfer remain a challenging problem in reinforcement learning where a powerful learning agent easily overfits to the environment or the task it has encountered, leading to poor generalization to new ones~\cite{overfit,huang2017adversarial}.

In this paper, we propose a new approach to tackle this challenge. Our main idea is to learn a meta rule to synthesize policies whenever the agent encounters new environments or tasks. Concretely, the meta rule uses the embeddings of the environment and the task to compose a policy, which is parameterized as the linear combination of the policy basis. On the training data from seen pairs of environments and tasks, our algorithm learns the embeddings as well as the policy basis. For new environments or tasks, the agent learns the corresponding embeddings only while it holds the policy basis fixed. Since the embeddings are low-dimensional, a limited amount of training data in the new environment or task is often adequate to learn well so as to compose the desired policy.  

While deep reinforcement learning algorithms are capable of memorizing and thus entangling representations of tasks and environments~\cite{overfit}, we propose a disentanglement objective such that the embeddings for the tasks and the environments can be extracted to maximize the efficacy of the synthesized policy. Empirical studies demonstrate the importance of disentangling the representations.

We evaluated our approach on \gridworld which we have created and the photo-realistic robotic environment \thor~\cite{Kolve2017AI2THORAI}. We compare to several leading methods for transfer learning in a significant number of settings. The proposed approach outperforms most of them noticeably in improving the effectiveness of transfer and adaptation. 

\section{Related Work}
\label{sRelated}

Multi-task~\cite{wilson2007multi} and transfer learning~\cite{Taylor2009TransferLF} for reinforcement learning (RL) have been long and extensively studied. Teh~\etal~\cite{distral} presented a distillation based method that transfers the knowledge from task specific agents to a multi-task learning agent. Andreas~\etal~\cite{Andreas2017ModularMR} combined the option framework~\cite{option} and modular network~\cite{module}, and presented an efficient multi-task learning approach which shares sub-policies across policy sketches of different tasks. Schaul~\etal~\cite{Schaul2015UniversalVF} encoded the goal state into value functions and showed its generalization to new goals. More recently, Oh~\etal~\cite{oh2017zero} proposed to learn a meta controller along with a set of parameterized policies to compose a policy that generalizes to unseen instructions. In contrast, we jointly consider the tasks and environments which can be both atomic, as we learn their embeddings without resorting to any external knowledge (e.g., text, attributes, etc.).

Several recent works~\cite{Dayan1993ImprovingGF,zhu2017visual,Barreto2017SuccessorFF,Kulkarni2016DeepSR} factorize  Q value functions  with an environment-agnostic state-action feature encoding function and task-specific embeddings. Our model is  related to this line of work in spirit. However, as opposed to learning the value functions, we  directly learn a factorized policy network with strengthened disentanglement between environments and tasks. This allows us to easily generalize better to new environments or tasks, as shown in the empirical studies.

\section{Approach}
\label{sApproach}

We begin by introducing notations and stating the research problem formally. We then describe the main idea behind our approach, followed by the details of each component of the approach.

\subsection{Problem Statement and Main Idea}

\paragraph{Problem statement.} We follow the standard framework for reinforcement learning~\cite{Sutton1998ReinforcementLA}.  An agent interacts with an environment by sequentially choosing actions over time and aims to maximize its cumulative rewards. This learning process is abstractly described by a Markov decision process with the following components:  a space of the agent's state $ s \in \mathcal{S}$, a space of possible actions $a \in \mathcal{A}$, an initial distribution of states $p_0(s)$, a stationary  distribution characterizing how the state at time $t$ transitions to the next state at $(t+1)$:   $p(s_{t+1}|s_{t},a_{t})$, and a reward function $r := r(s, a)$. 

The agent's actions follow a policy $\pi(a|s): \mathcal{S}\times\mathcal{A}\rightarrow [0,1]$, defined as a conditional distribution $p(a|s)$. The goal of the learning is to identify the optimal policy that maximizes the discounted cumulative reward: $R = \mathbb{E}[\sum_{t=0}^{\infty}\gamma^{t}r(s_{t},a_{t})]$, where $\gamma\in (0,1]$ is a discount factor and the expectation is taken with respect to the randomness in state transitions and taking actions. We denote by $p(s| s',t,\pi)$ the probability at state $s$ after transitioning  $t$ time steps, starting from  state $s'$ and following the policy $\pi$. With it, we define the discounted state distribution as $\rho^{\pi}(s)=\sum_{s'}\sum_{t=1}^{\infty}\gamma^{t-1} p_{0}(s')  p(s|s', t, \pi)$.

In this paper, we study how an agent learns to accomplish a variety of tasks in different environments. Let $\mathcal{E}$ and $\mathcal{T}$ denote the sets of the environments and the tasks, respectively. We assume the cases of finite sets but it is possible to extend our approach to infinite ones. While the most basic approach is to learn an optimal policy under each pair $(\varepsilon, \tau)$ of environment and task, we are interested in \emph{generalizing to all combinations in $(\calE, \calT)$,  with interactive learning from a limited subset of $(\varepsilon, \tau)$ pairs}. Clearly, the smaller the subset is, the more desirable the agent's generalization capability is.

\paragraph{Main idea.} In the rest of the paper, we refers to the limited subset of pairs as \emph{seen pairs} or \emph{training pairs} and the rest ones as \emph{unseen pairs} or \emph{testing pairs}. We assume that the agent does \emph{not} have access to the unseen pairs to obtain any interaction data to learn the optimal policies directly.  In computer vision, such problems have been intensively studied in the frameworks of unsupervised domain adaptation and zero-shot learning, for example, \cite{Gong2012GeodesicFK,Changpinyo2016SynthesizedCF,Chao2016AnES,Misra2017FromRW}. There are totally $|\calE|\times |\calT|$ pairs -- our goal is to learn from $O(|\calE| +  |\calT|)$ training pairs and generalize to all.


Our main idea is to synthesize policies for the unseen pairs of environments and tasks. In particular, our agent learns two sets of embeddings: one for the environments and the other for the tasks. Moreover, the agent also learns how to compose policies using such embeddings. Note that learning both the embeddings and how to compose happens on the training pairs. For the unseen pairs, the policies are constructed and used right away --- if there is interaction data, the policies can be further fine-tuned. However, even without such interaction data, the synthesized policies still perform well. 

To this end, we desire our approach to jointly supply two aspects: a compositional structure of \textbf{Synthesized Policies (\ourmethod)} from environment and task embeddings and a disentanglement learning objective to learn the embeddings. We refer this entire framework as \ourmethod and describe its details in what follows.

\begin{figure}[t]
	\centering
	\includegraphics[width=0.975\textwidth]{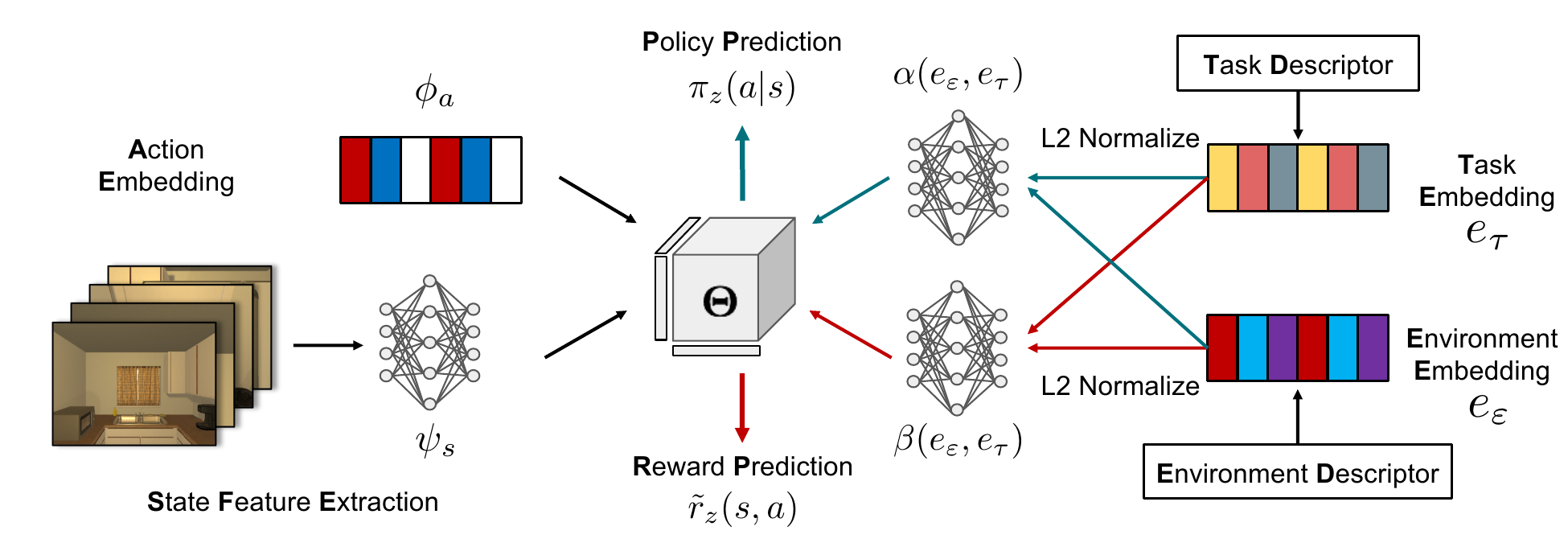}
	\caption{\footnotesize Overview of our proposed model. Given a  task and an environment, the corresponding embeddings $e_\varepsilon$ and $e_\tau$ are retrieved to compose the policy coefficients and reward coefficients. Such coefficients then linearly combine the shared basis and synthesize a policy (and a reward prediction) for the agent.}
	\label{fig:overview}
\end{figure}

\subsection{Policy Factorization and Composition}

Given a pair $z=(\varepsilon, \tau)$ of an environment $\varepsilon$ and a task $\tau$, we denote by $e_\varepsilon$ and $e_\tau$ their embeddings, respectively. The policy is synthesized with a bilinear mapping
\begin{equation}
\label{eqn:sythesize_policy}
\pi_z (a|s) \propto \exp(\vpsi_s\T \mU(e_\varepsilon, e_\tau) \vphi_a + b_\pi)
\end{equation}
where $b_\pi$ is a scalar bias, and $\vpsi_s$ and $\vphi_a$ are featurized states and actions (for instances, image pixels or the feature representations of an image). The bilinear mapping given by the matrix $\mU$ is parameterized as the linear combination of  $K$ basis matrices $\mTheta_k$,
\begin{equation}
\mU(e_\varepsilon, e_\tau) = \sum_{k=1}^K \alpha_k (e_\varepsilon, e_\tau) \mTheta_k. 
\end{equation}
Note that the combination coefficients depend on the specific pair of environment and task while the basis is \emph{shared} across all pairs. They enable knowledge transfer from the seen pairs to unseen ones.

Analogously, during learning (to be explained in detail  in the later section),  we predict  the rewards by modeling them with the same set of basis but different combination coefficients:
\begin{equation}
\label{eqn:sythesize_reward}
\tilde{r}_z(s, a) = \vpsi_s\T \mV(e_\varepsilon, e_\tau) \vphi_a + b_r = \vpsi_s\T \left(\sum_k \beta_k (e_\varepsilon, e_\tau) \mTheta_k \right)\vphi_a + b_r
\end{equation}
where $b_r$ is a scalar bias.
Note that similar strategies for learning to predict rewards along with learning the policies  have also been studied in recent works~\cite{Jaderberg2016ReinforcementLW,zhu2017visual,Barreto2017SuccessorFF}. We find this strategy helpful too (cf.\ details in our empirical studies in Section~4).

Figure~\ref{fig:overview} illustrates the model architecture described above. In this paper, we consider agents that take egocentric views of the environment, so a convolutional neural network is used to extract the state features $\vpsi_s$ (cf.\ the bottom left panel of Figure~\ref{fig:overview}). The action features $\vphi_a$ are learned as a look-up table. Other model parameters include the basis $\mTheta$, the embeddings $e_\varepsilon$ and $e_\tau$ in the look-up tables respectively for the environments and the tasks, and the coefficient functions $\alpha_k(\cdot, \cdot)$ and $\beta_k(\cdot, \cdot)$ for respectively synthesizing the policy and reward predictor. 
The coefficient functions $\alpha_k(\cdot, \cdot)$ and $\beta_k(\cdot, \cdot)$ are parameterized with one-hidden-layer MLPs with the inputs being the concatenation of  $e_\varepsilon$ and $e_\tau$, respectfully.

\subsection{Disentanglement of the Embeddings for Environments and Tasks}

In \ourmethod, both the embeddings and the bilinear mapping are to be learnt. In an alternative but equivalent form, the policies are formulated as
\begin{equation}
{\pi}_z(a|s) \propto \exp\left( \sum_k \alpha_k (e_\varepsilon, e_\tau)  \vpsi_s\T  \mTheta_k  \vphi_a + b_{\pi}\right).
\end{equation}
As the defining coefficients $\alpha_k$ are parameterized by a neural network whose inputs and parameters are both optimized, we need to impose additional structures such that the learned embeddings facilitate the transfer across environments or tasks. Otherwise, the learning could overfit to the seen pairs and consider each pair in unity, thus leading to poor generalization to unseen pairs.

To this end, we introduce discriminative losses to distinguish different environments or tasks through the agent's trajectories.  Let $\vx = \{\vpsi_s\T  \mTheta_k  \vphi_a\} \in \R^K$ be the state-action representation. 
For the agent interacting with an environment-task pair $z = (\varepsilon, \tau)$, we denote its trajectory as $\{\vx_1, \vx_2, \cdots, \vx_t, \ldots\}$. 
We argue that a good embedding (either $e_\varepsilon$ or $e_\tau$) ought to be able to tell from which environment or task the trajectory is from. In particular, we formulate this as a multi-way classification where we desire $\vx_t$ (on average) is telltale of its environment $\varepsilon$ or task $\tau$:
\begin{align}
\ell_\varepsilon := - \sum_t \log P(\varepsilon | \vx_t) & \text{ with } P(\varepsilon | \vx_t) \propto \exp\left( g(\vx_t)\T e_\varepsilon \right) \\
\ell_\tau     := - \sum_t \log P(\tau | \vx_t)   & \text{ with } P(\tau | \vx_t)\propto \exp\left( h(\vx_t)\T e_\tau \right)
\end{align}
where we use two nonlinear mapping functions ($g(\cdot)$ and $h(\cdot)$, parameterized by one-hidden-layer MLPs) to transform the state-action representation $\vx_t$, such that it retrieves $e_\varepsilon$ and $e_\tau$. These two functions are also learnt using the interaction data from the seen pairs.

\subsection{Learning}
\label{sec:learning}

Our approach (\ourmethod) relies on the modeling assumption that the policies (and the reward predicting functions) are factorized in the axes of the environment and the task. This is a generic assumption and can be integrated with many reinforcement learning algorithms. In this paper, we study its effectiveness on imitation learning (mostly) and also reinforcement learning.  

In imitation learning, we denote by $\pi_z^e$ the expert policy of combination $z$ and apply the simple strategy of ``behavior cloning'' with random perturbations to learn our model from the expert demonstration~\cite{ho2016generative}. 
We employ a cross-entropy loss for the policy as follows:
\[
\ell_{\pi_z} := - \mathbb{E}_{s\sim\rho^{\pi_z^e}, a \sim \pi_z^e}[ \log\pi_{z}(a|s) ]
\]
A $\ell_2$ loss is used for learning the reward prediction function, $\ell_{r_z} := \mathbb{E}_{s\sim\rho^{\pi_z^e}, a \sim \pi_z^e}\lVert \tilde{r}_{z}(s,a) - {r}_{z}(s, a) \rVert_2$.
Together with the disentanglement losses, they form the overall loss function 
\[
\mathcal{L}:=\mathbb{E}_{z}[\ell_{\pi_z} + \lambda_1 \ell_{r_z} + \lambda_2 \ell_{\varepsilon} + \lambda_3 \ell_{\tau}]
\]
which is then optimized through experience replay, as shown in \textbf{Algorithm 1} in the supplementary materials (Suppl. Materials). We choose the value of those hyper-parameters $\lambda_i$ so that the contributions of the objectives are balanced. More details are presented in the Suppl. Materials.

\subsection{Transfer to Unseen Environments and Tasks}
\label{sec:transfer}

Eq.~\ref{eqn:sythesize_policy} is used to synthesize a policy for any (\env, \task) pair, as long as the environment and the task --- not necessarily the pair of them --- have appeared at least once in the training pairs. If, however, a new environment and/or a new task appears (corresponding to the transfer setting 2 or 3 in Section~\ref{sIntro}), fine-tuning is required to extract their embeddings. To do so, we keep all the components of our model fixed except the look-up tables (i.e., embeddings) for the environment and/or the task. This effectively re-uses the policy composition rule and enables  fast learning of the environment and/or the task embeddings, after seeing a few number of demonstrations. In the experiments, we find it works well even with only one shot of the demonstration. 

\section{Experiments}
\label{sExp}
We validate our approach (\ourmethod) with extensive experimental studies, comparing with several baselines and state-of-the-art transfer learning methods.

\subsection{Setup}
We experiment with two simulated environments\footnote{\scriptsize The implementation of \ourmethod and the gridworld environment are available on \url{https://www.github.com/sha-lab/SynPo} and \url{https://www.github.com/sha-lab/gridworld}, respectfully.}: \gridworld and \thor~\cite{Kolve2017AI2THORAI}, in both of which the agent takes as input an egocentric view (cf.\ Figure~\ref{fig:simulators}). Please refer to the Suppl.\ Materials for more details about the state feature function $\vpsi_s$ used in these simulators.

\begin{figure}[t]
	\centering
	\begin{tabular}{ccc}
		\includegraphics[width=0.61\textwidth]{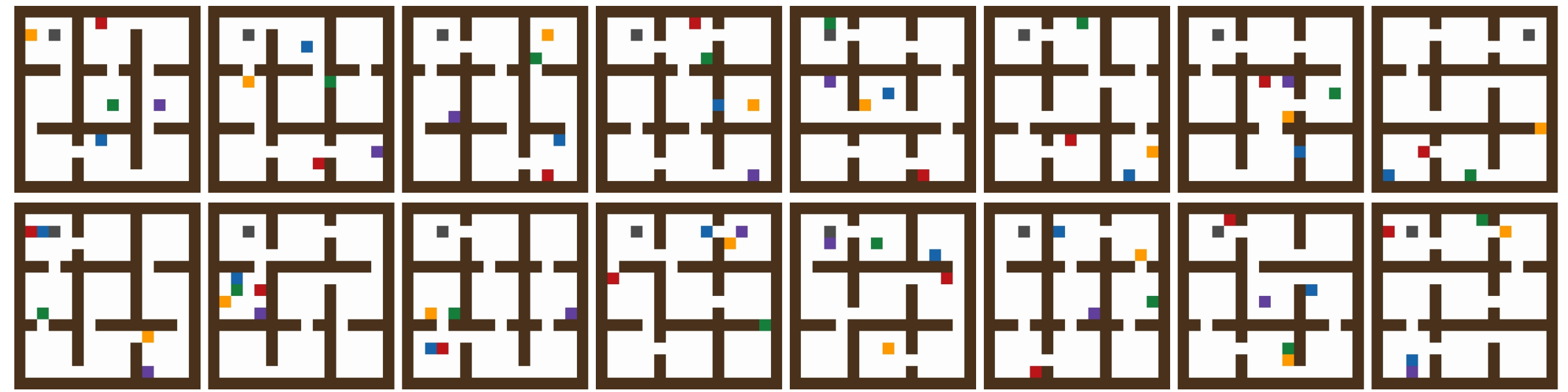} &
		\includegraphics[width=0.15\textwidth]{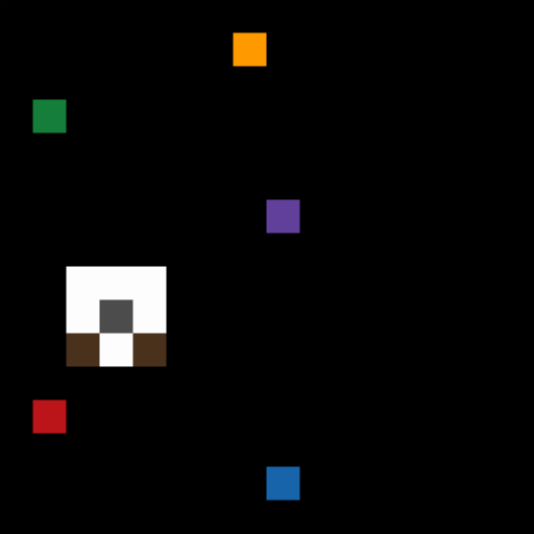} &
		\includegraphics[width=0.15\textwidth]{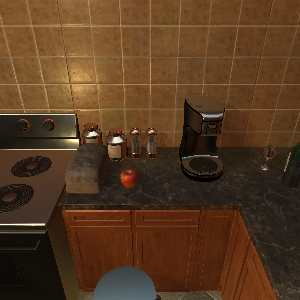}
	\end{tabular}
	
	\caption{\small From left to right: (a) Some sample mazes of our \gridworld dataset. They are similar in appearance but different in topology.
		Demonstrations of an agent's egocentric views of (b) \gridworld  and (c) \thor.}
	\label{fig:simulators}
\end{figure}

\paragraph{\gridworld and tasks.} We design twenty $16\times16$ grid-aligned mazes,  some of which are visualized in Figure~\ref{fig:simulators} (a). The mazes are similar in appearance but differ from each other in topology. There are five colored blocks as ``treasures'' and the agent's goal is to collect the treasures in pre-specified orders, \eg, \textit{``Pick up Red and then pick up Blue''}. At a time step, the ``egocentric'' view observed by the agent consists of  the agent's surrounding within a $3\times 3$ window and the treasures' locations. At each run, the locations of the agent and treasures are randomized. We consider twenty tasks in each environment, resulting $|\calE| \times |\calT| = 400$ pairs of (\env, \task) in total. In the transfer setting 1 (cf. Figure~\ref{fig:illustration}(a)), we randomly choose 144 pairs as the training set under the constraint that each of the environments appears at least once, so does any task. The remaining 256 pairs are used for testing. For the transfer settings 2 and 3 (cf.\ Figure~\ref{fig:illustration}(b) and (c)), we postpone the detailed setups to Section~\ref{sec:exp:transfer_unseen}. 

\paragraph{\thor\cite{Kolve2017AI2THORAI} and tasks.} We also test our method on \thor, a challenging 3D simulator where the agent is placed in indoor photo-realistic scenes. The tasks are to search and act on objects, \eg, \textit{``Put the cabbage to the fridge''}. Different from \gridworld, the objects' locations are unknown so the agent has to search for the objects of interest by its understanding of  the visual scene (cf.\ Figure~\ref{fig:simulators}(c)). There are 7 actions in total (\textit{look up}, \textit{look down}, \textit{turn left}, \textit{turn right}, \textit{move forward}, \textit{open/close}, \textit{pick~up/put down}). We run experiments with 19  scenes $\times$ 21 tasks in this simulator. 

\paragraph{Evaluations.} We evaluate the agent's performance by the averaged success rate (AvgSR.) for accomplishing the tasks, limiting the maximum trajectory length to 300 steps. For the results reported in numbers (\eg, Tables~\ref{tab:gridworld}), we run 100 rounds of experiments for each (\env, \task) pair by randomizing the agent's starting point and the treasures' locations. To plot the convergence curves (\eg, Figure~\ref{fig:success_rate}), we sample 100 (\env, \task) combinations and run one round of experiment for each to save computation time. We train our algorithms under 3 random seeds and report the mean and standard deviation (std).

\paragraph{Competing methods.} We compare our approach (\ourmethod ) with the following baselines and competing methods. Note that our problem setup is new, so we have to adapt the competing methods, which were proposed for other scenarios, to fit ours. 

\begin{itemize}[leftmargin=*]
	\item \textbf{MLP.} The policy network is a multilayer perceptron whose input concatenates state features and the environment  and task embeddings. We train this baseline using the proposed losses for our approach, including the disentanglement losses $\ell_\epsilon, \ell_\tau$; it performs worse without $\ell_\epsilon, \ell_\tau$.
	\item \textbf{Successor Feature (SF).} We learn the \textit{successor feature} model~\cite{Barreto2017SuccessorFF} by Q-imitation learning for fair comparison. {We strictly follow \cite{Kulkarni2016DeepSR} to set up the learning objectives}. The key difference of SF from our approach is its lack of capability in capturing the environmental priors.
	\item \textbf{Module Network (ModuleNet).} We also implement a module network following~\cite{devin2017learning}. Here we train an environment specific module for each environment and a task specific module for each task. The policy for a certain (\env, \task) pair is assembled by combining the corresponding environment module and task module.
	\item \textbf{Multi-Task Reinforcement Learning (MTL).} This is a degenerated version of our method, where we ignore the distinctions of environments. We simply replace the environment embeddings by zeros for the coefficient functions. The disentanglement loss on task embeddings is still used since it leads to better performances than otherwise.
\end{itemize}

Please refer to the Suppl. Materials for more experimental details, including all the twenty \gridworld mazes, how we configure the rewards, optimization techniques, feature extraction for the states, and our implementation of the baseline methods.

\begin{figure}[t]
	\centering
	\begin{tabular}{cc}
		\includegraphics[width=0.475\textwidth,trim={1cm 0 2.cm 1.5cm},clip]{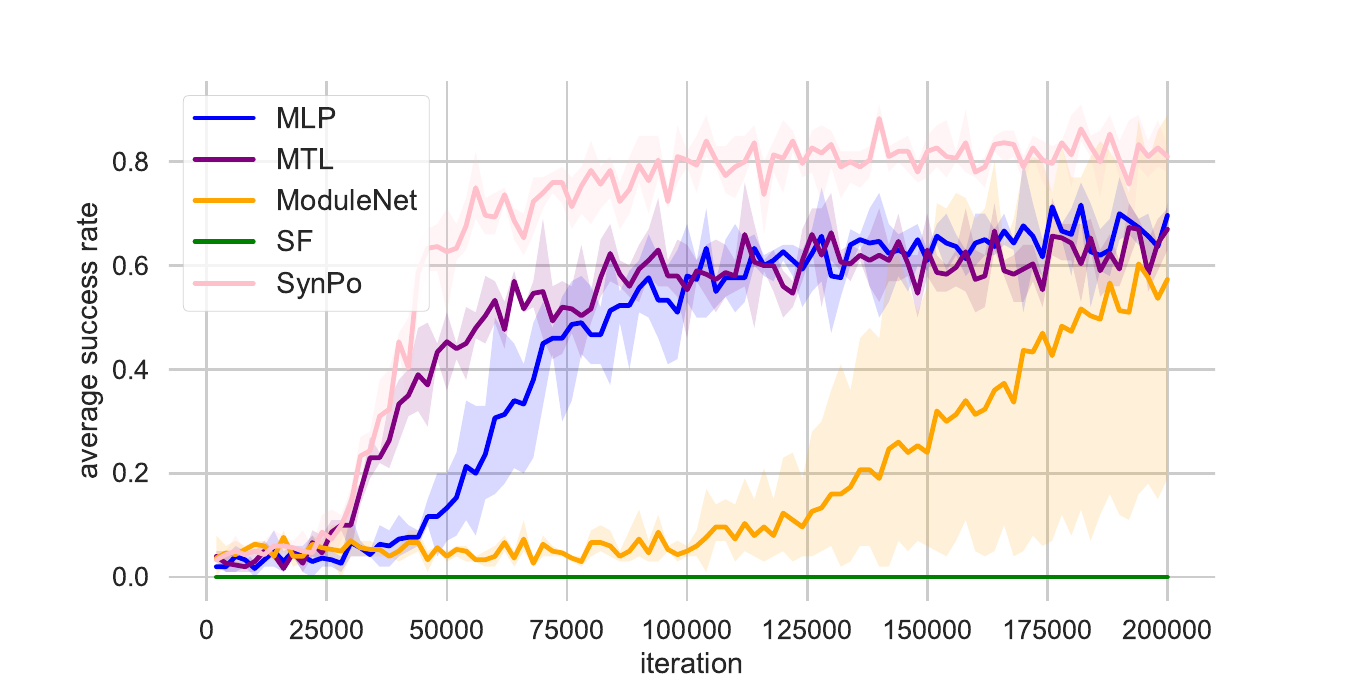} &
		\includegraphics[width=0.475\textwidth,trim={1cm 0 2.cm 1.5cm},clip]{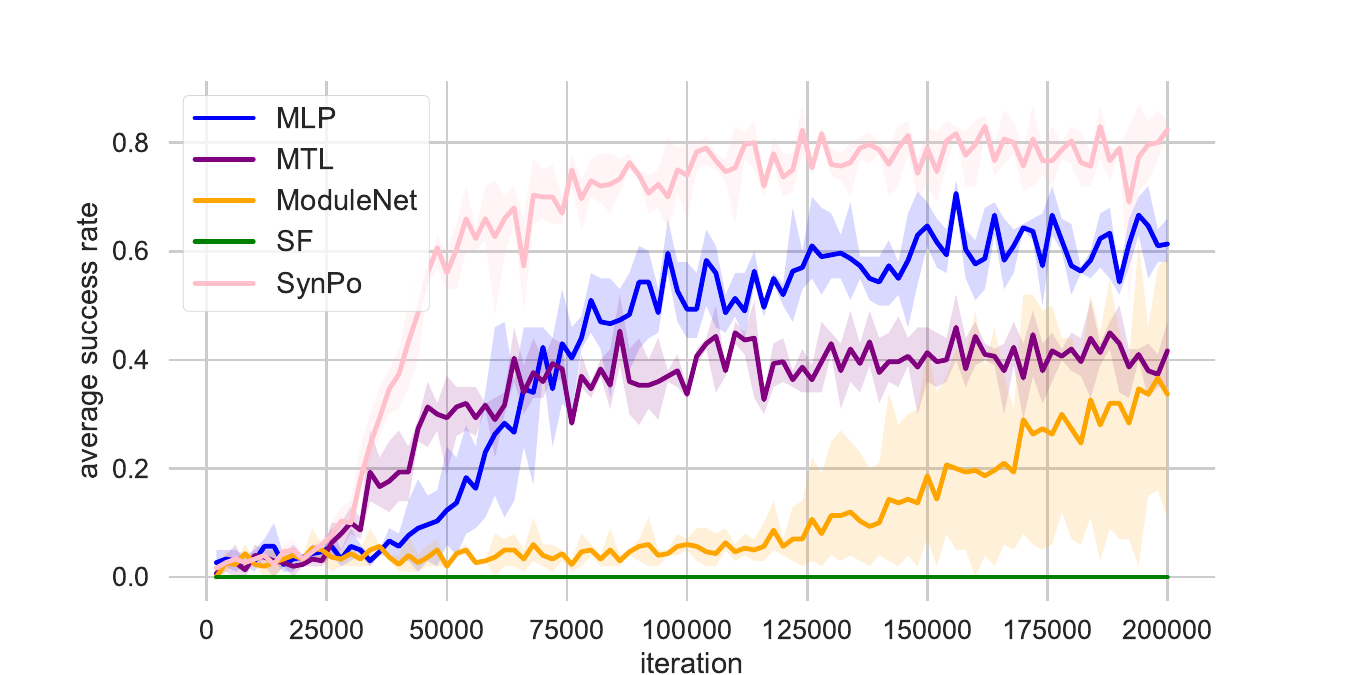} \\
		(a) AvgSR. over Time on \textsc{seen} & (b) AvgSR. over Time on \textsc{unseen} \\
	\end{tabular}
	\caption{\small \textbf{On \gridworld.} Averaged success rate (AvgSR) on \textsc{seen} pairs and \textsc{unseen} pairs, respectively. Results are reported with \textit{$|\calE| = 20$ and $|\calT| = 20$}. We report mean and std based on 3 training random seeds. }
	\label{fig:success_rate}
\end{figure}

\begin{figure}[t]
	\centering
	\begin{tabular}{cc}
		\includegraphics[width=0.475\textwidth,trim={1cm 0cm 2.5cm 1.5cm},clip]{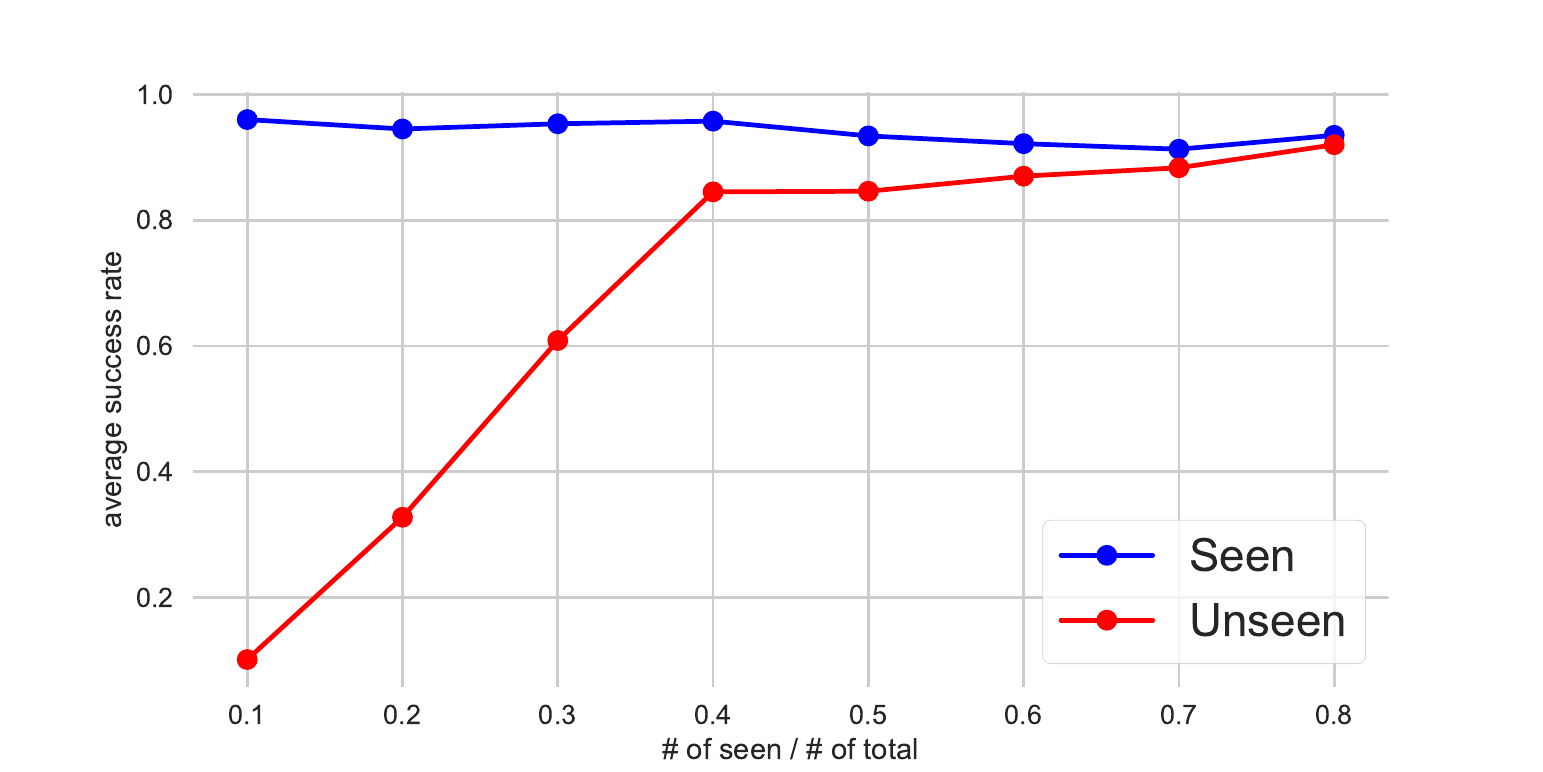} &
		\includegraphics[width=0.475\textwidth,trim={1cm 0 2.cm 0.8cm},clip]{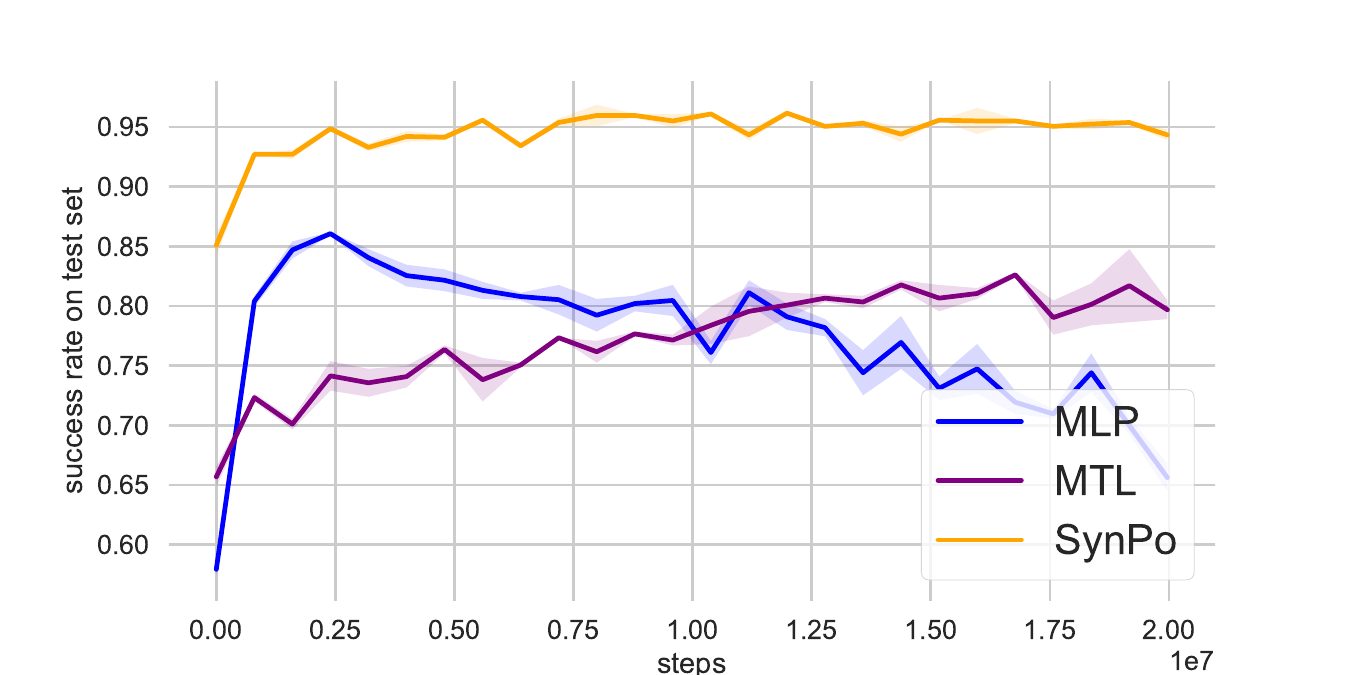} \\
		(a) Transfer learning performance curve & (b) AvgSR. over Time on \textsc{unseen} \\
	\end{tabular}
	\caption{\small (a) \textbf{Transfer learning performance} (in AvgSR.) with respect to the ratio: \# \textsc{Seen} pairs / \# \textsc{Total} pairs, with \textit{$|\calE| = 10$ and $|\calT| = 10$}. (b) \textbf{Reinforcement learning performance} on unseen pairs of different approaches (with PPO~\cite{schulman2017proximal}). MLP overfits, MTL improves slightly, and \ourmethod achieves 96.16\% AvgSR.} 
	\label{fig:transfer_rl}
\end{figure}

\subsection{Experimental Results on \gridworld}

We first report results on the adaptation and transfer learning setting 1, as described in Section 1 and Figure~\ref{fig:illustration}(a). There, the agent acts upon a new pair of environment and task, both of which it has encountered  during training but not in the same  (\env, \task) pair. The goal is to use as sparse   (\env, \task) pairs among all the combinations as possible to learn and yet still able to transfer successfully. 

\subsubsection{Transfer to Previously Encountered Environments and Tasks}

\begin{table}[t]
	\centering
	\setlength{\tabcolsep}{5pt}
	\caption{Performance (AvgSR.) of each method on \gridworld (\textsc{Seen}/\textsc{Unseen} = 144/256). 
	}
	\begin{tabular}{c | cccc | c}
		Method & SF & ModuleNet & MLP & MTL & \ourmethod \\ \hlineB{3}
		AvgSR. (\textsc{Seen}) & 0.0 $\pm$ 0.0\% & 50.9 $\pm$ 33.8\% & 69.0 $\pm$ 2.0\% & 64.1 $\pm$ 1.2\% & \textbf{83.3 $\pm$ 0.5 \%} \\
		AvgSR. (\textsc{Unseen}) & 0.0 $\pm$ 0.0\% & 30.4 $\pm$ 20.1\% & 66.1 $\pm$ 2.6\% & 41.5 $\pm$ 1.4\% & \textbf{82.1 $\pm$ 1.5\%} \\
	\end{tabular}
	\label{tab:gridworld}
\end{table}

\begin{table}[t]
	\caption{Performance of transfer learning in the settings 2 and 3 on \gridworld}
	\label{tab:extension}
	\centering
	\begin{tabular}{cc|c|c|c}
		Setting & Method & Cross Pair ($Q$'s \env, $P$'s \task)  & Cross Pair ($P$'s \env, $Q$'s \task) & $Q$ Pairs\\ \hlineB{3}
		\multirow{2}{*}{Setting 2} & MLP &  13.8\% & 20.7\% & 6.3\% \\
		& \ourmethod & \textbf{50.5\%} & \textbf{21.5\%} & \textbf{13.5\%} \\ \hline
		\multirow{2}{*}{Setting 3} & MLP &  14.6\% & 18.3\% & 7.2\% \\
		& \ourmethod &   \textbf{42.7\%} & \textbf{19.4\%} & \textbf{12.9\%} \\  
	\end{tabular}
\end{table}

\paragraph{Main results.} Table~\ref{tab:gridworld} and Figure~\ref{fig:success_rate} show the success rates and convergence curves, respectively, of our approach and the competing methods averaged over the seen and unseen (\env, \task) pairs. 
\ourmethod consistently outperforms the others in terms of both the convergence and final performance, by a significant margin. 
On the seen split, MTL and MLP have similar performances, while MTL performs worse comparing to MLP on the unseen split (\ie in terms of the generalization performance), possibly because it treats all the environments the same. 

We design an extreme scenario to further challenge the environment-agnostic methods (e.g., MTL). We reduce the window size of the agent's view to one, so the agent sees the cell it resides and the treasures' locations and nothing else. As a result, MTL suffers severely, MLP performs moderately well, and \ourmethod outperforms both significantly \textit{(unseen AvgSR: MTL=6.1\%, MLP=66.1\%, \ourmethod = 76.8\%)}. We conjecture that the environment information embodied in the states is crucial for the agent to beware of and generalize across distinct environments. More discussions are deferred to the Suppl.\ Materials. 

\paragraph{How many seen (\env, \task) pairs do we need to transfer well?} Figure~\ref{fig:transfer_rl}(a) shows that, not surprisingly, the transfer learning performance increases as the number of seen pairs increases. The acceleration slows down after the seen/total ratio reaches 0.4. In other words, when there is a limited budget, our approach enables the agent to learn from 40\% of all possible (\env, \task) pairs and yet generalize well across the tasks and environments. 

\paragraph{Does reinforcement learning help transfer?} Beyond imitation learning, we further study our {\ourmethod} for reinforcement learning (RL) under the same transfer learning setting. Specifically, we use PPO~\cite{schulman2017proximal} to fine-tune the three top performing algorithms on \gridworld. The results averaged over 3 random seeds are shown in Figure~\ref{fig:transfer_rl}(b). We find that RL fine-tuning improves the transfer performance for all the three algorithms. In general, MLP suffers from over-fitting, MTL is improved moderately yet with a significant gap to the best result, and \ourmethod achieves the best AvgSR, \textbf{96.16\%}.

\paragraph{Ablation studies.} We refer readers to the Suppl. Materials for ablation studies of the learning objectives. 

\subsubsection{Transfer to Previously Unseen Environments or Tasks}
\label{sec:exp:transfer_unseen}

Now we investigate how effectively one can schedule transfer from seen environments and tasks to unseen ones, i.e., the settings 2 and 3 described in Section~\ref{sIntro} and Figure~\ref{fig:illustration}(b) and (c). The seen pairs (denoted by $P$) are constructed from ten environments and ten tasks; the remaining ten environments and ten tasks are unseen (denoted by $Q$). Then we have two settings of transfer learning. 



One is to transfer to pairs which cross the seen set $P$ and unseen set $Q$ -- this corresponds to the setting 2 as the embeddings for either the unseen tasks or the unseen environments need to be learnt, but not both. Once these embeddings are learnt, we use them to synthesize policies for the test (\env, \task) pairs. This mimics the style \textit{``incremental learning of small pieces and integrating knowledge later''}.

The other is the transfer setting 3. The agent learns policies via learning embeddings for the tasks and environments of the unseen set $Q$ and then composing, as described in section 3.5. Using the embeddings from $P$ and $Q$, we can synthesize policies for any  (\env, \task) pair.  This mimics the style of \textit{``learning in giant jumps and connecting dots''}.


\paragraph{Main results.} Table~\ref{tab:extension} contrasts the results of the two transfer learning settings. Clearly, setting 2 attains stronger performance as it ``incrementally learns'' the embeddings of either the tasks or the environments but not both, while setting 3 requires learning both simultaneously. It is interesting to see this result aligns with how effective human learns. 

Figure~\ref{fig:extensibility} visualizes the results whose rows are indexed by tasks and columns by environments. The seen pairs in {$P$ are in the upper-left quadrant} and the unseen set {$Q$ is on the bottom-right}. We refer readers to the Suppl.\ Materials for more details and discussions of the results.

\begin{figure}[ht]
	\centering
	\begin{tabular}{cc}
		\includegraphics[width=0.48\textwidth]{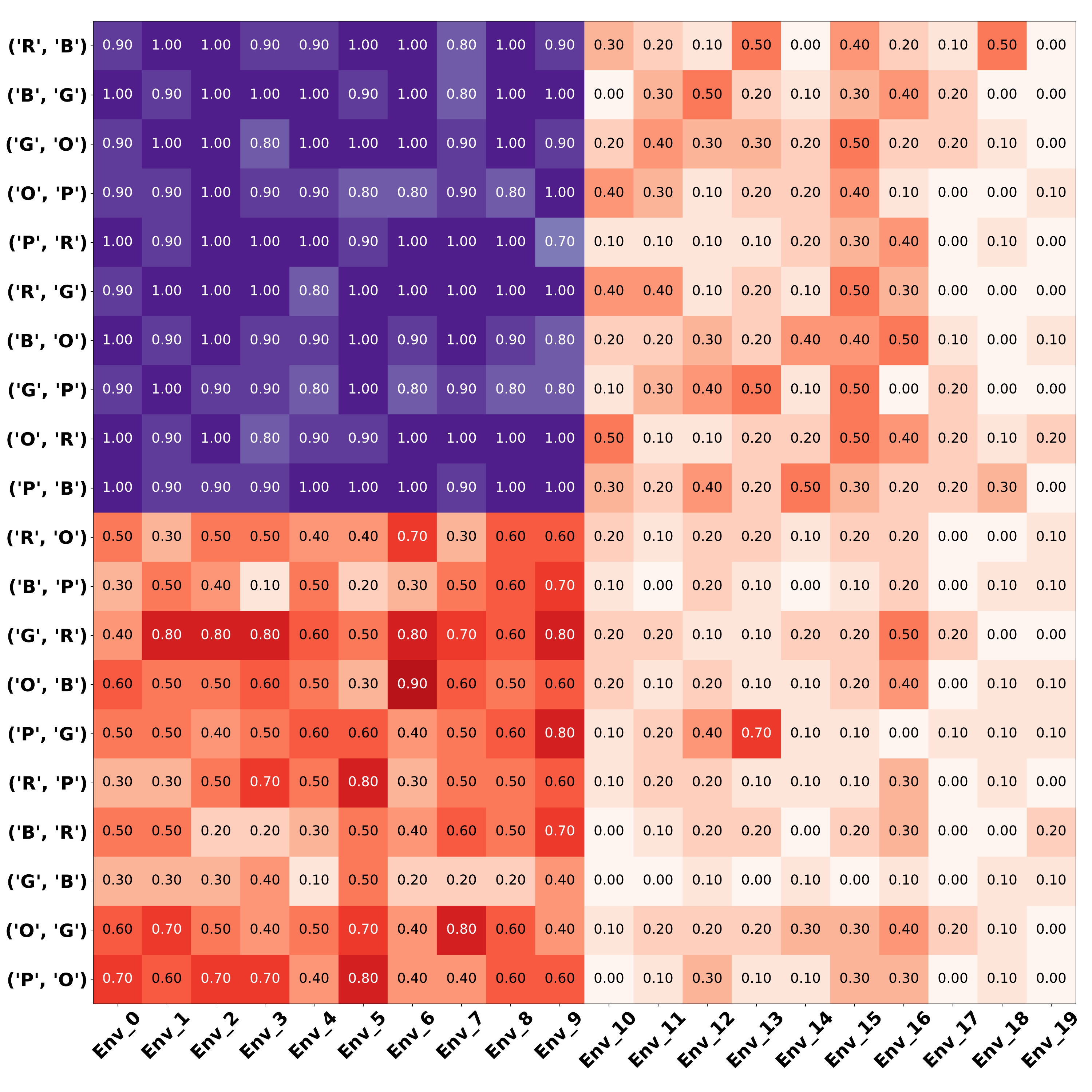} & 
		\includegraphics[width=0.48\textwidth]{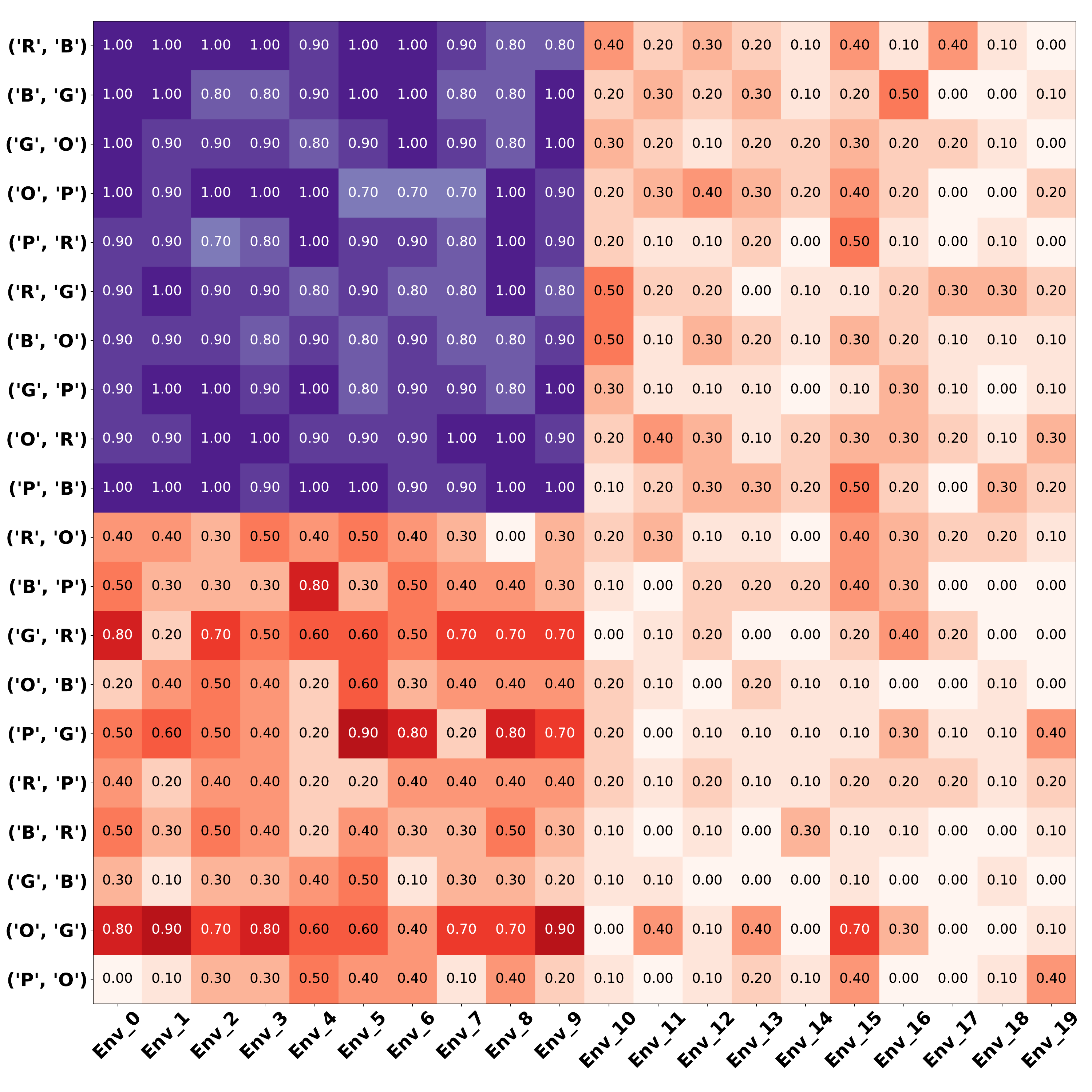} \\
		(a) Transfer Setting 2 & (b) Transfer Setting 3
	\end{tabular}
	\caption{\small Transfer results of settings 2 and 3. AvgSRs are marked in the grid (see Suppl.\ Materials for more visually discernible plots). The tasks and environments in the \textcolor{blue}{\textbf{purple cells}} are from the unseen $Q$ set and the \textcolor{red}{\textbf{red cells}} correspond to the rest. Darker color means  better performance. It shows that cross-task transfer is easier than cross-environment.}
	\label{fig:extensibility}
	\vskip-1em
\end{figure}

\subsection{Experimental Results on \thor}

\begin{table}[t]
	\centering
	\setlength{\tabcolsep}{15pt}
	\caption{Performance of  each method on \thor (\textsc{Seen}/\textsc{Unseen}=144/199)}
	\begin{tabular}{c| ccc |c}
		Method & ModuleNet& MLP & MTL & \ourmethod \\ \hlineB{3} 
		AvgSR. (\textsc{Seen})   & 51.5 \% & 47.5\% & 52.2\% & \textbf{55.6\%} \\
		AvgSR. (\textsc{Unseen}) & 14.4 \% & 25.8\% & 33.3\% & \textbf{35.4\%} \\
	\end{tabular}
	\label{tab:thor}
\end{table}

\paragraph{Main results.} The results on the \thor simulator are shown in Table~\ref{tab:thor}, where we report our approach as well as the top performing ones on \gridworld. Our {\ourmethod} significantly outperforms three competing ones for both seen pairs and unseen pairs. Moreover, our approach also has the best performance of success rate on seen to unseen, indicating that it is less prone to overfiting than the other methods. More details are included in the Suppl.\ Materials.

\section{Conclusion}
\label{sConclusion}

In this paper, we consider the problem of learning to simultaneously transfer across both environments~(\env) and tasks~(\task) under the reinforcement learning framework and, more importantly, by learning from only sparse (\env, \task) pairs out of all the possible combinations. Specifically, we present a novel approach that learns to synthesize policies from the disentangled embeddings of environments and tasks. We evaluate our approach for the challenging transfer scenarios in two simulators, \gridworld and \thor. Empirical results verify that our method generalizes better across environments and tasks than several competing baselines.

{\small \noindent \textbf{Acknowledgments} We appreciate the feedback from the reviewers. This work is partially supported by DARPA\# FA8750-18-2-0117, NSF IIS-1065243, 1451412, 1513966/ 1632803/1833137, 1208500, CCF-1139148, a Google Research Award, an Alfred P. Sloan Research Fellowship, gifts from Facebook and Netflix, and ARO\# W911NF-12-1-0241 and W911NF-15-1-0484.}

{\footnotesize
	\bibliographystyle{ieee}
	\bibliography{rl}
}

\clearpage

\appendix

\begin{center}
	\Large\textbf{Supplementary Material}
\end{center}

In this Supplementary Material, we provide details omitted in the main paper:

\begin{itemize}[leftmargin=*,noitemsep]
	\item Section~\ref{sec:details_simulator}: Detailed configurations about \gridworld and \thor simulators.
	\item Section~\ref{sec:imitation}: Imitation learning algorithm and optimization details.
	\item Section~\ref{sec:reinforce}: Reinforcement learning algorithm and optimization details.
	\item Section~\ref{sec:details_implement}: Implementation details about SynPo and baselines.
	\item Section~\ref{sec:additional_exp}: Additional experimental results to the main text.
\end{itemize}

\section{Details on simulators}
\label{sec:details_simulator}

\subsection{Details about \gridworld Configurations} 

As we have mentioned in the main text, there are in total 20 environments for this simulator, which we listed as Figure~\ref{fig:gridworld_maps}. The tasks presented in this simulator includes a sequential execution of picking up two treasures in different colors. The agent can observe the layout of the environment inside a 3x3 square centered at the agent's current position (see Figure~\ref{fig:simulators} (a) for details). The agent can take 5 actions, which includes moving in the four directions and picking up an object right below it. Note that in each run of a certain given task, the locations of both agent and treasures are randomized.

In terms of the reward setting, we follow the common practice and set the reward for moving one step to be -0.01 and touching a wall to be an additional - 0.01. Picking up a target treasure gives 1 unit of the reward and completing a task gives 10 unites of the reward. Picking up a wrong target directly ends an episode and gives reward -10. During the training, we use an optimal planner with shortest path search algorithm for expert policy. To represent a state for our network, we follow the practice in DQN~\cite{Mnih2015HumanlevelCT} and concatenate the last four observations as the input to the policy. 

\begin{figure}[h]
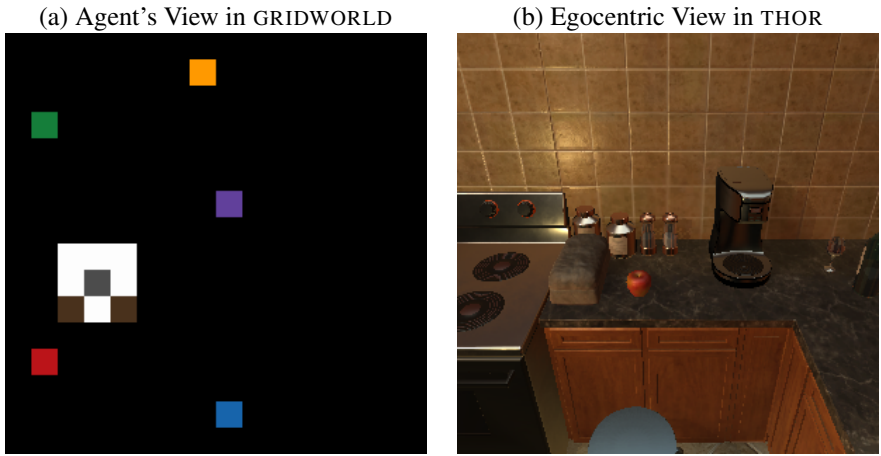

	\centering
	\begin{tabular}{cc}
		(a) Agent's View in \gridworld & (b) Egocentric View in \thor \\
		\includegraphics[width=0.4\textwidth]{./figures/env-illustration-gridworld-agent} &  \includegraphics[width=0.4\textwidth]{./figures/env-illustration-thor} \\
	\end{tabular}
	\caption{Demonstrations of agent's view in two simulators. In the left, we present the agent's input state of \gridworld. An agent only have the vision to its surrounding context and the locations of all treasures (see (a)). Similarly, in the \thor, an agent has access to an egocentric image that represents the first-person viewpoint  (see (b)).}
	\label{fig:simulators}
\end{figure}

\subsection{Details about \thor Configurations.} 

\thor~\cite{Kolve2017AI2THORAI} is a 3D robotic simulator developed recently for simulating the indoor environments a robot could encounter. An agent is working like a real robot with a first-persion view camera, which delivers RGB images in egocentric view (see Figure~\ref{fig:simulators} (b) for details). The environment has interactable components that a agent can play with, which enables the learning of human like behaviors such as semantic planning~\cite{zhu2017visual} and indoor navigation~\cite{Zhu2017TargetdrivenVN}. We describe the concrete settings we used as what follows. 

We extract the image features using convolutional neural networks to represent an observation for each egocentric view of a robotic agent. Specifically, we extract the activation output from the penultimate layer of a Resnet101~\cite{resnet} pre-trained on ImageNet~\cite{imagenet}, which has the dimensionality of 2048. Similar to the \gridworld experiments, we then concatenate those features of the last four observations as the input to the policy network. The agent can take 7 actions in \thor: move ahead, turn left, turn right, look up, look down, open/close an object, pick up/put down an object. We set the reward for moving one step to be -0.01 and executing invalid actions to be -0.01. The reward of picking up the correct object is 1, and the reward of finishing the task is 10. Picking up the wrong object and putting the object in the wrong receptacle ends an episode and gives -10 units of the reward. The interactable objects, receptacles and index of environments (kitchens) are listed in table~\ref{tab:interactable}. In our experiment we selected environments with similar size (see Table~\ref{tab:interactable} for the complete list).

\begin{table}[h]
	\centering
	\setlength{\tabcolsep}{10pt}
	\caption{interactable objects, receptacles and environment indexes in \thor}
	\begin{tabular}{l|c}
		Entries & Values \\
		\hlineB{3}
		Objects & Container, Lettuce, Mug, Tomato, Plate, Apple, Bowl \\
		\hline
		Receptacles & Fridge, Microwave, Sink \\
		\hline
		Environments & Kitchen \{1, 2, 3, 4, 5, 6, 8, 9, 11, 12, 18, 22, 23, 24, 25, 27, 28, 20, 30\}
	\end{tabular}
	\label{tab:interactable}
\end{table}

\section{Imitation Learning Algorithm and Optimization Details}
\label{sec:imitation}

As mentioned in the main text, now we describe the imitation learning algorithm used for learning \ourmethod and all baseline models. The concrete details are presented in Algorithm~\ref{alg_imitation}. 

\begin{algorithm}[ht]
	\caption{Policy Imitation Learning Algorithm.}
	\label{alg_imitation}
	\begin{algorithmic}
		\STATE {\bfseries Input:} Given training simulators $\mathbf{simulator}({z})$, where $z \in (\calE, \calT)_{\texttt{train}}$
		\STATE Initialize Expert Replay Memory $\calD^{\mathbf{E}}$ with the capacity $\mathbf{N}$
		\FOR{episode = 1, M}
		\STATE Sample $z \in (\calE, \calT)_{\texttt{train}}$
		\STATE $\textsc{traj}_{z}\xspace(\{ s_i, a_i, r_i\};{\pi^{\mathbf{E}}})$ = $\rollout\Big(\pi^{\mathbf{E}}_{z}, \mathbf{simulator}({z})\Big)$
		\STATE Store $\textsc{traj}_{z}\xspace(\{ s_i, a_i, r_i\};{\pi^{\mathbf{E}}})$ to $\calD^{\mathbf{E}}$
		\STATE Sample a random mini-batch $\mathbf{B}$ with $| \, \mathbf{B} \, |$ trajectories from $\calD^{\mathbf{E}}$
		\STATE Compute gradient $\nabla \calL$ and update the parameters with specified optimizer
		\ENDFOR
	\end{algorithmic}
\end{algorithm}

In each episode, we sample a trajectory using the expert policy and store it into the replay buffer.  At the end of each episode, we sample 64 trajectories uniformly from the replay buffer to calculate the total loss. Here, the size of replay buffer for storing expert trajectories is 20,000. In each episode, we uniformly sample 64 trajectories from the replay buffer (coming from different \env and \task pairs) to compute the loss. We set the hyper-parameters $\lambda$ as follows:  there is $\lambda_{1}=0.01$ for reward prediction; $\lambda_{2} = 0.1$ and $\lambda_{3} = 0.001$ for environment and task disentanglement loss. The dimensionality of environment embedding and task embedding are 128. Besides, we use Adam~\cite{adam} as the optimizer with the initial learning rate set to be 0.001. Additionally, we set the value of weight decay factor to be $0.001$ in all our experiments. 

\section{Reinforcement Learning Algorithm and Optimization Details}
\label{sec:reinforce}

As mentioned in the main text, we have employed reinforcement learning to further fine-tune our model, which archived improvement in transfer learning performances. Now we describe the detailed setups of our experiments. We use PPO~\cite{schulman2017proximal} to fine-tune our model. We optimize our model by RMSProp with learning rate $0.000025$ and weight decay $0.0001$. We use GAE~\cite{schulman2015high} to calculate advantages, with $\gamma=0.99$ and $\lambda=0.95$, entropy weight is $0.01$, rollout length $128$, objective clipping ratio $0.1$. Gradient norms are clipped to $0.5$. We divide the trajectories collected into 4 mini batches and do four optimization steps on each update. We fine-tuned our model for $2 \times 10^7$ steps. During RL fine-tuning we also included our disentangling objectives as auxilary loss.

\section{Implementation Details}
\label{sec:details_implement}

\subsection{Details about our Policy Network for \ourmethod in \gridworld}

First, we introduce the specific setups we used for policy networks in \gridworld. We directly parameterize the outcome of a dot product between $\mTheta$ and $\phi_a$ as a tensor, for the sake of computation efficiency in practice. However, our model, as mentioned in the main text, is indeed a bilinear policy. Therefore, with a more general application scenario that action space ($|\calA|$) is large, we can apply the original form of our approach and learn separate action embeddings $\phi_a$ with the shared basis $\mTheta$. The coefficient functions $\alpha(\cdot)$ and $\beta(\cdot)$ that compose environment and task embeddings are one-hidden-layer MLPs with 512 hidden units and output size of 128. The dimension of the state feature $\psi_s$ extracted from ResNet before the bilinear weight $\mU$ is 128. The state feature extractor is a customized ResNet. Its concrete structure is shown as Table~\ref{tab:resnet}. The dimensionality of the environment embeddings $e_\varepsilon$ and task embeddings $e_\tau$ are 128.

\begin{table}[h]
	\centering
	\caption{Structure of State Feature Function $\psi_s$ in \gridworld}
	\begin{tabular}{c|c|cc}
		\hline
		group name & output size & block type & stride \\[2pt]
		\hline
		input  & $16 \times 16 \times 3$   & - & - \\[4pt]
		conv 1 & $8 \times  8  \times 32$  & $\left[ {\begin{array}{c}
			3\times 3, 32   \\
			3\times 3, 32   \\
			\end{array}} \right] \times 2$ & 2 \\[8pt]
		conv 2 & $4 \times  4 \times 64$  & $\left[ {\begin{array}{c}
			3\times 3, 64   \\
			3\times 3, 64   \\
			\end{array}} \right] \times 2$ & 2 \\[8pt]
		conv 3 & $2 \times  2  \times 128$ & $\left[ {\begin{array}{c} 3\times 3, 128   \\
			3\times 3, 128  
			\end{array}} \right] \times 2$ & 2 \\[8pt]
		conv 4 & $2 \times  2  \times 256$ & $\left[ {\begin{array}{c} 3\times 3, 256   \\ 
			3\times 3, 256  
			\end{array}} \right] \times 2$ & 0 \\[8pt]
		fc     & 128 & $\Big[ 1024 \times 128 \Big] $ & - \\[4pt] \hline 
	\end{tabular}
	\label{tab:resnet}
\end{table}

\subsection{Details about our Policy Network for \ourmethod in \thor}

Next we describe the network setups we used in \thor. Again, we directly parameterize the outcome of a dot product between $\mTheta$ and $\phi_a$ as a tensor, as the action space is small ($|\calA| = 7$) in this simulator. With the stacked $2,048 \times 4$ dimensional ResNet101 feature as input, we learn a two 1-D  convolutional networks with kernel size of 3 and stride of 2, which first reduces the dimensionality of feature to 1,024 and then aggregates over the temporal axis. Next, the encoding of visual feature is then concatenated with an embedding ($e_{\mathbf{obj}}$) that represents object the agent is carrying. The concatenated feature vector is next input into a one-hidden-layer MLP wth hidden state of 2,048 dimension. The output of this MLP (which is also the final output ofstate feature function $\psi_s$) has dimension of 256. The concrete config is shown as Table~\ref{tab:tc}. The dimensionality of the environment embeddings $e_\varepsilon$ and task embeddings $e_\tau$ are 128.

\begin{table}[h]
	\centering
	\caption{Structure of State Feature Function $\psi_s$ in \thor}
	\begin{tabular}{c|c|cc}
		\hline
		group name & output size & block type & stride  \\[2pt]
		\hline
		image input  & $2048 \times 4$ & - & -  \\[6pt]
		conv 1 & $1024 \times 2$ & $\Big[ 3\times 1, 1024 \Big]$ & 2 \\[8pt]
		conv 2 & $1024 \times 1$ & $\Big[ 3\times 1, 1024 \Big]$ & 2 \\[8pt] \hline
		concat & 1056 & concat $e_{\mathbf{obj}}$ & - \\[4pt] \hline 
		fc1     & 2048 & $\Big[ 1056 \times 2048 \Big]$ & - \\[8pt]
		fc2     & 256  & $\Big[ 2048 \times 256 \Big]$  & - \\[8pt] \hline
	\end{tabular}
	\label{tab:tc}
\end{table}

\subsection{Details about learning Disentanglement Objective}

In addition to both of the above settings, we applied another set of one-hidden-layer MLPs $f_\varepsilon$ and $f_\tau$ (hidden=512) to represent the auxiliary function that project the high-dimensional trajectory feature $\vx$ to the embedding spaces $e_\varepsilon$ and $e_\tau$. Note that this function is only used in the disentanglement objective, and could be discarded during the deployment of policy network.

\section{Additional Experimental Results}
\label{sec:additional_exp}

\subsection{Complete Details of Main Results and Comparison between Methods}

As mentioned in the main text, we put our complete results of \gridworld here. Now we report not only the average success rate (AvgSR.) but also average reward (AvgReward), on both {seen} and {unseen} pairs. 

\begin{table}[h]
	\centering
	\setlength{\tabcolsep}{5pt}
	\caption{Performance of the best model for each method on \gridworld ({Seen}/{Unseen}=144/256). All algorithms are trained using three random seeds and reported with mean and std. on each (\env, \task) pair, we sample the locations of agent and treasures for 100 times to evaluate the performances.}
	\begin{tabular}{c | cccc | c}
		Method & SF & ModuleNet & MLP & MTL & \ourmethod \\ \hlineB{3}
		AvgSR. (\textsc{Seen}) & 0.0 $\pm$ 0.0\% & 50.9 $\pm$ 33.8\% & 69.0 $\pm$ 2.0\% & 64.1 $\pm$ 1.2\% & \textbf{83.3 $\pm$ 0.5 \%} \\
		AvgSR. (\textsc{Unseen}) & 0.0 $\pm$ 0.0\% & 30.4 $\pm$ 20.1\% & 66.1 $\pm$ 2.6\% & 41.5 $\pm$ 1.4\% & \textbf{82.1 $\pm$ 1.5\%} \\
	\end{tabular}
	\label{tab:gridworld}
\end{table}

\begin{figure}[h]
	\centering
	\begin{tabular}{cc}
		\includegraphics[width=0.475\textwidth,trim={2.25cm 0 2cm 1.5cm},clip]{figures/gridworld/w1/overall_success_train_w1-ny} &
		\includegraphics[width=0.475\textwidth,trim={2.25cm 0 2cm 1.5cm},clip]{figures/gridworld/w1/overall_success_test_w1-ny} \\
		(a) AvgSR. over Time on \textsc{seen} & (b) AvgSR. over Time on \textsc{unseen} \\
		\includegraphics[width=0.475\textwidth,trim={2.25cm 0 2cm 1.5cm},clip]{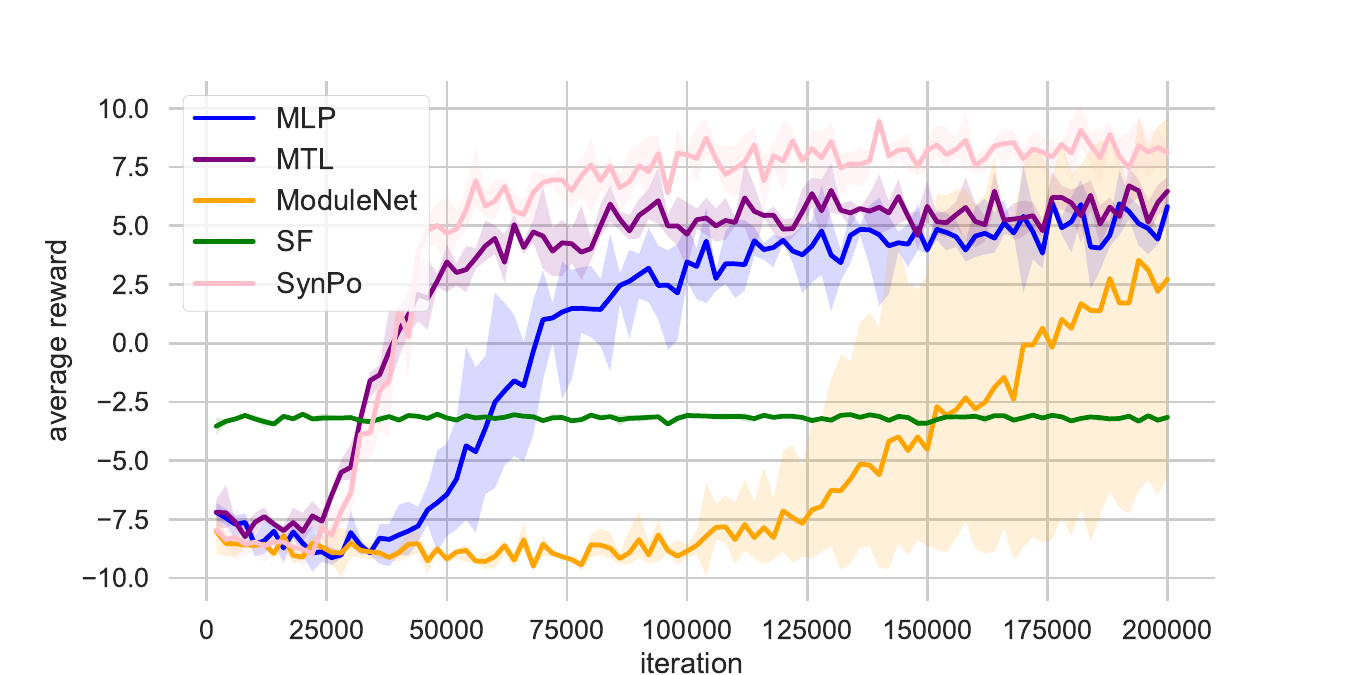} &
		\includegraphics[width=0.475\textwidth,trim={2.25cm 0 2cm 1.5cm},clip]{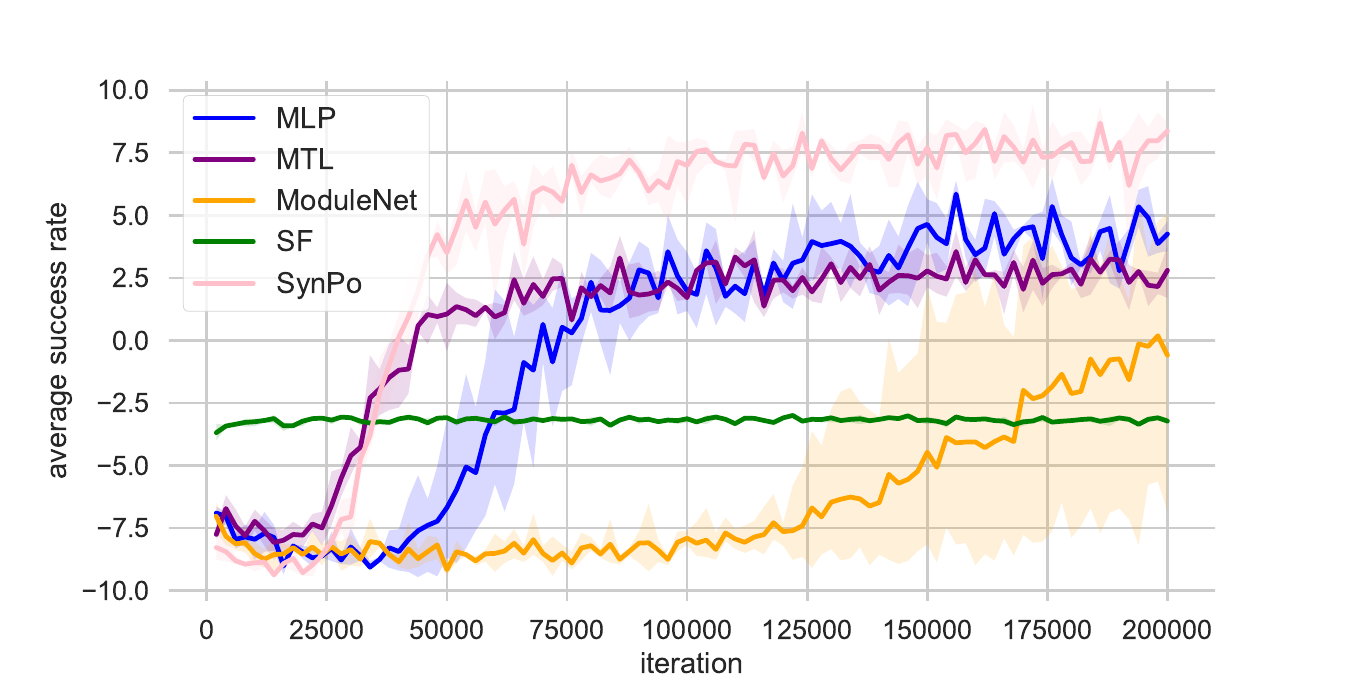} \\
		(c) AvgReward over Time on \textsc{seen} & (d) AvgReward over Time on \textsc{unseen} \\
	\end{tabular}
	\caption{\textbf{Results on \gridworld.} (a)-(b): Comparison between average success rate (ASR.) of algorithms on {seen} split and {unseen} split. (c)-(d): Comparison between average accumulated reward (AvgReward.) of algorithms in each episode on {seen} split and {unseen} split. Results are reported on the setting with \textit{$|\calE| = 20$ and $|\calT| = 20$}. For each intermediate performance, we sample 100 (\env, \task) combinations and test one configuration to evaluate the performances. We evaluate models trained with 3 random seeds and report results in terms of the mean AvgSR and its standard deviation. }
	\label{fig:success_rate}
\end{figure}

\begin{figure}[t!]
	\centering
	\begin{tabular}{cc}
		\includegraphics[width=0.475\textwidth,trim={2.25cm 0 2cm 1.5cm},clip]{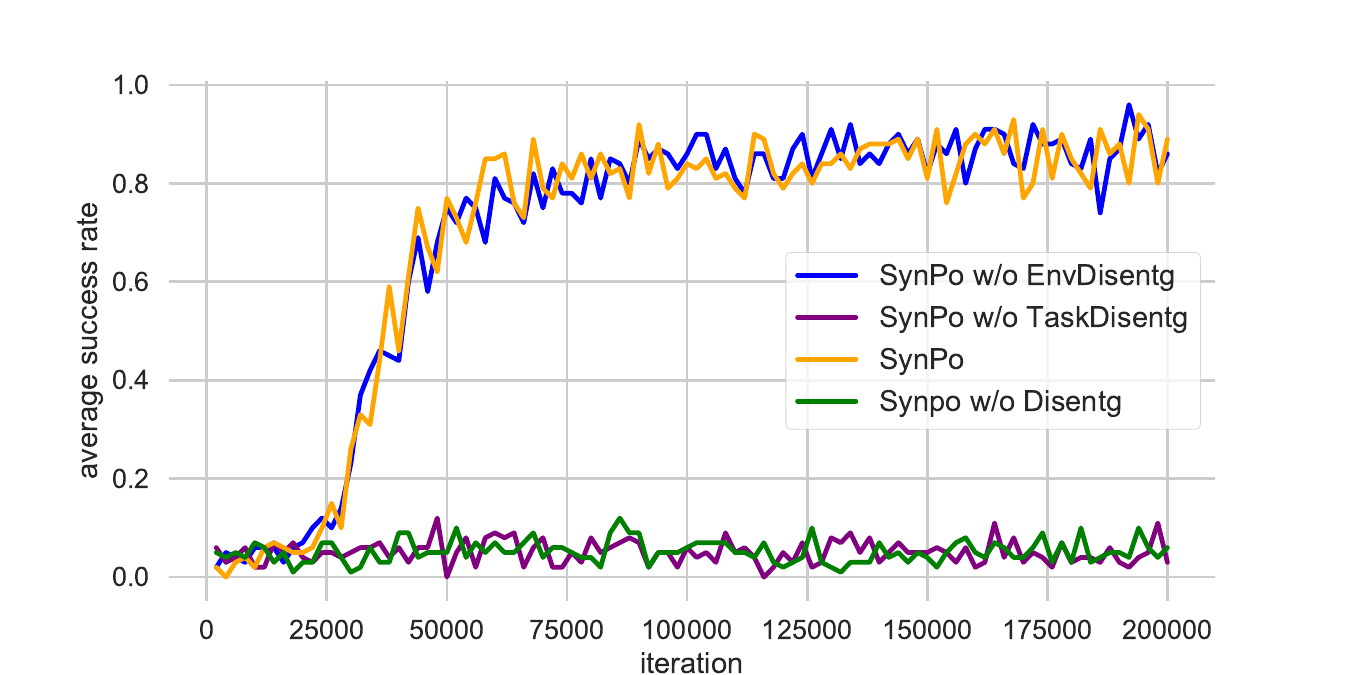} &
		\includegraphics[width=0.475\textwidth,trim={2.25cm 0 2cm 1.5cm},clip]{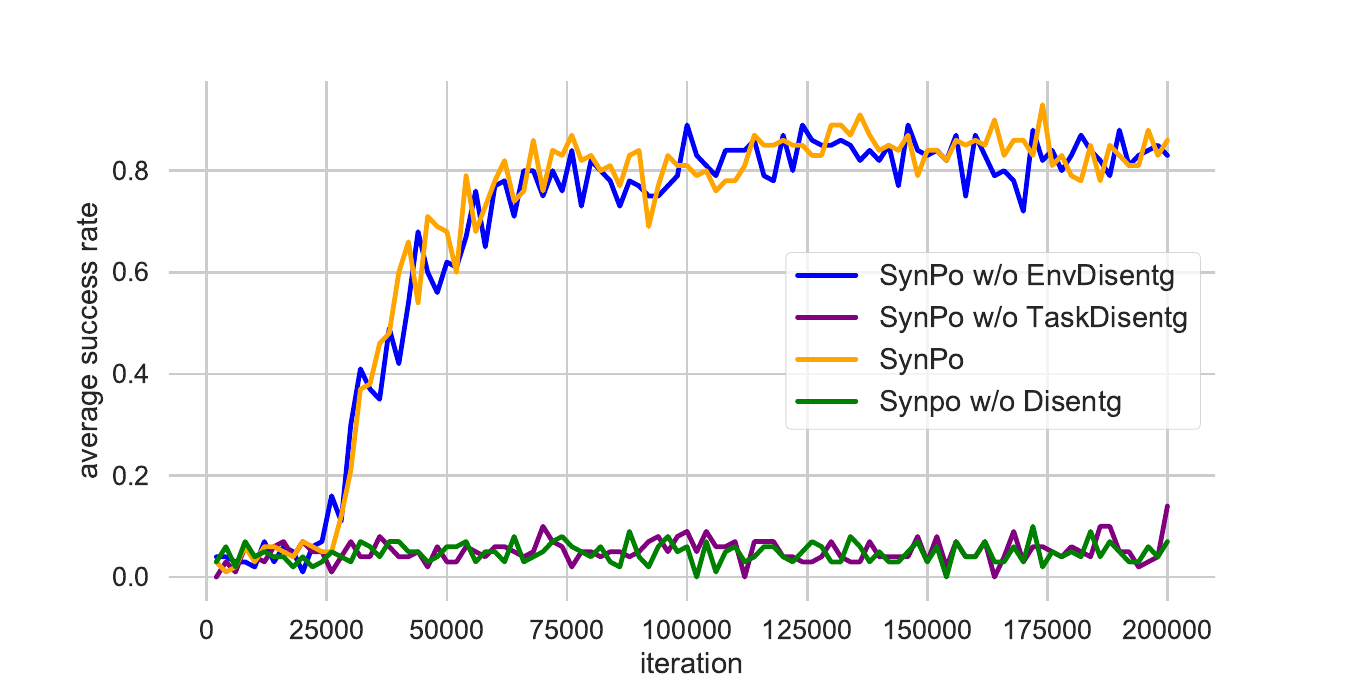} \\
		(a) AvgSR. over Time on \textsc{seen} & (b) AvgSR. over Time on \textsc{unseen} \\
		\includegraphics[width=0.475\textwidth,trim={2.25cm 0 2cm 1.5cm},clip]{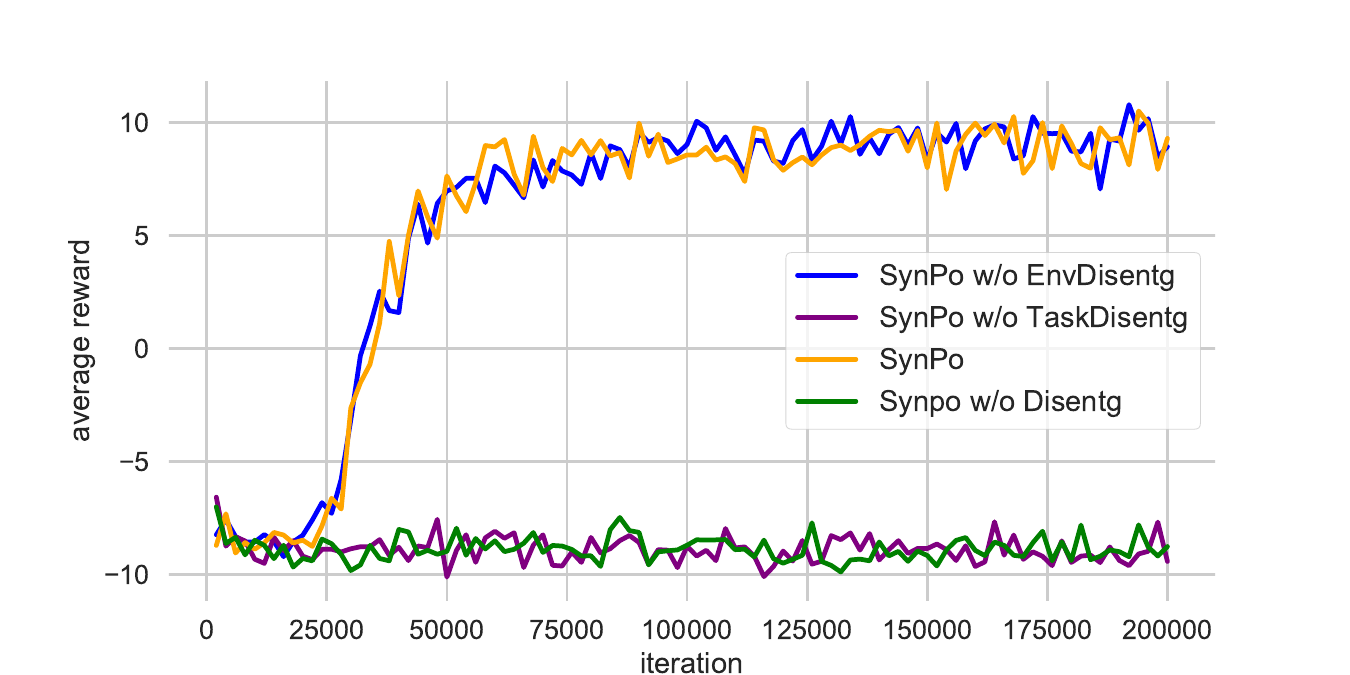} &
		\includegraphics[width=0.475\textwidth,trim={2.25cm 0 2cm 1.5cm},clip]{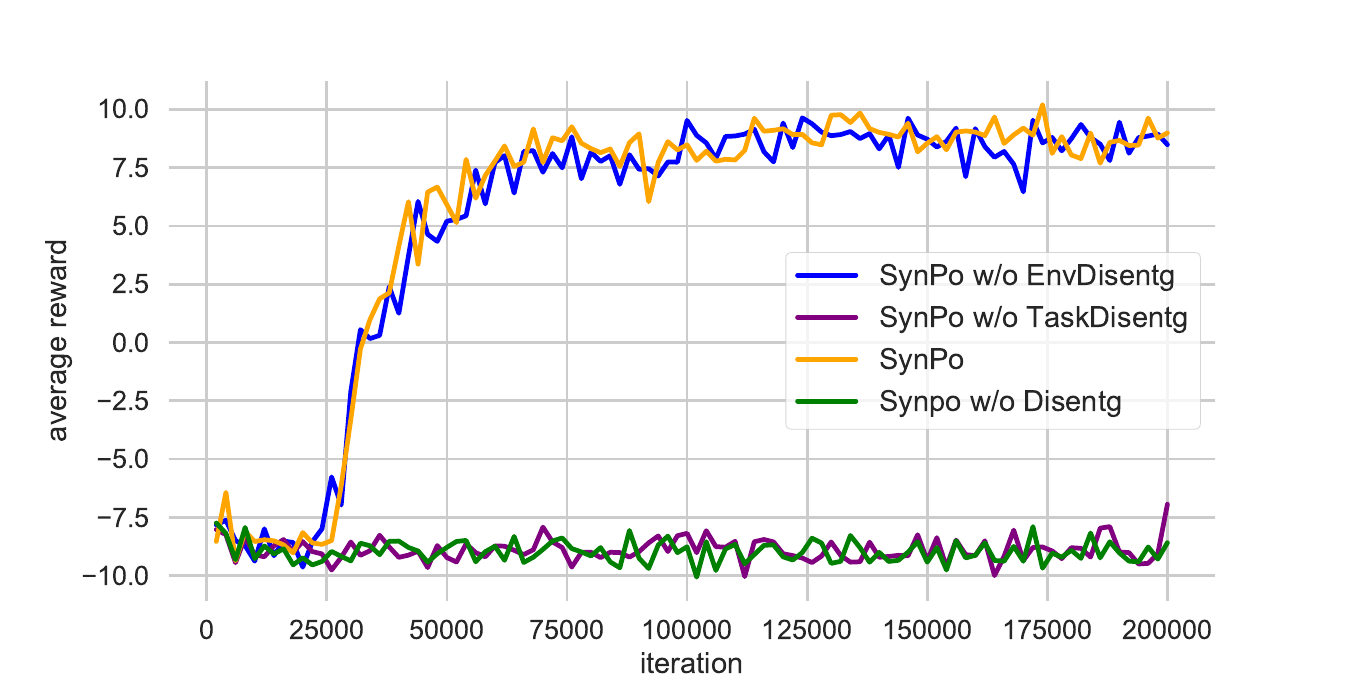} \\
		(c) AvgReward over Time on \textsc{seen} & (d) AvgReward over Time on \textsc{unseen} \\
	\end{tabular}
	\caption{\textbf{An ablation study about our learning objectives.} We report the results of the ablated versions without the disentanglement loss (Disentg) on environment (EnvDisentg) and on task (TaskDisentg). (a)-(b): Comparison between average success rate (ASR.) of algorithms on \textsc{seen} split and \textsc{unseen} split. (c)-(d): Comparison between average accumulated reward (AvgReward.) of algorithms in each episode on \textsc{seen} split and \textsc{unseen} split. Results are reported on the setting with \textit{$|\calE| = 20$ and $|\calT| = 20$}. Similarly, for each intermediate performance, we sample 100 (\env, \task) combinations to evaluate the performances.}
	\label{fig:objectives}
\end{figure}

We found that the trend of average reward on {seen} and {unseen} splits are quite similar to the trend of average success rate. We also note that the reward for successor feature (SF) is stable around -3, which indicated that the agent only tries to avoid negative reward and refuse to learn getting positive reward. On the contrary, all methods that make progress later starts with a lower average reward, meaning that the agent tries to complete the task by picking up objects but failed a lot at the beginning. 

Specifically, we find that \ourmethod is consistently performing better across all metrics, in terms of both the convergence and final performance. On the seen splits, MTL and MLP have similar performances, while MTL has a much worse generalization performance on unseen splits, comparing to MLP, possibly due to over-fitting or the lack of the capability in recognizing environments. At the same time, it is worth noting that Module Network has a significantly larger variance in its performances, comparing against all other approaches. This is possibly due to the fact that the environment modules and task modules are adhered together during the inference, where instability could occur. Similar issue has also been reported by Devin \etal~\cite{devin2017learning}. In addition, even in the best performing cases, ModuleNet could achieve a similar performances comparing to MLP and still far from approaching \ourmethod's performance. 

\subsection{Ablation Studies of the Learning Objectives}

How does each component in the objective function of our approach affect the performance of our model?  Figure~\ref{fig:objectives} shows that the task disentanglement loss is crucial for achieving good success rates on either seen or unseen pairs. This is probably because the differences between tasks are very subtle, making the agent hard to find the right distinct embeddings for them without the explicit task disentanglement loss. In contrast, the approach without the environment disentanglement loss can still reach a high success rate though it converges a bit slower.

\subsection{Details on Transfer Learning Experiments}

\begin{figure}[h]
	\small
	\centering
	\label{tab:confusion1}
	\begin{tabular}{cc}
		\textbf{(a) 10 Train and 90 Test } & \textbf{(b) 20 Train and 80 Test } \\
		\includegraphics[width=0.475\textwidth]{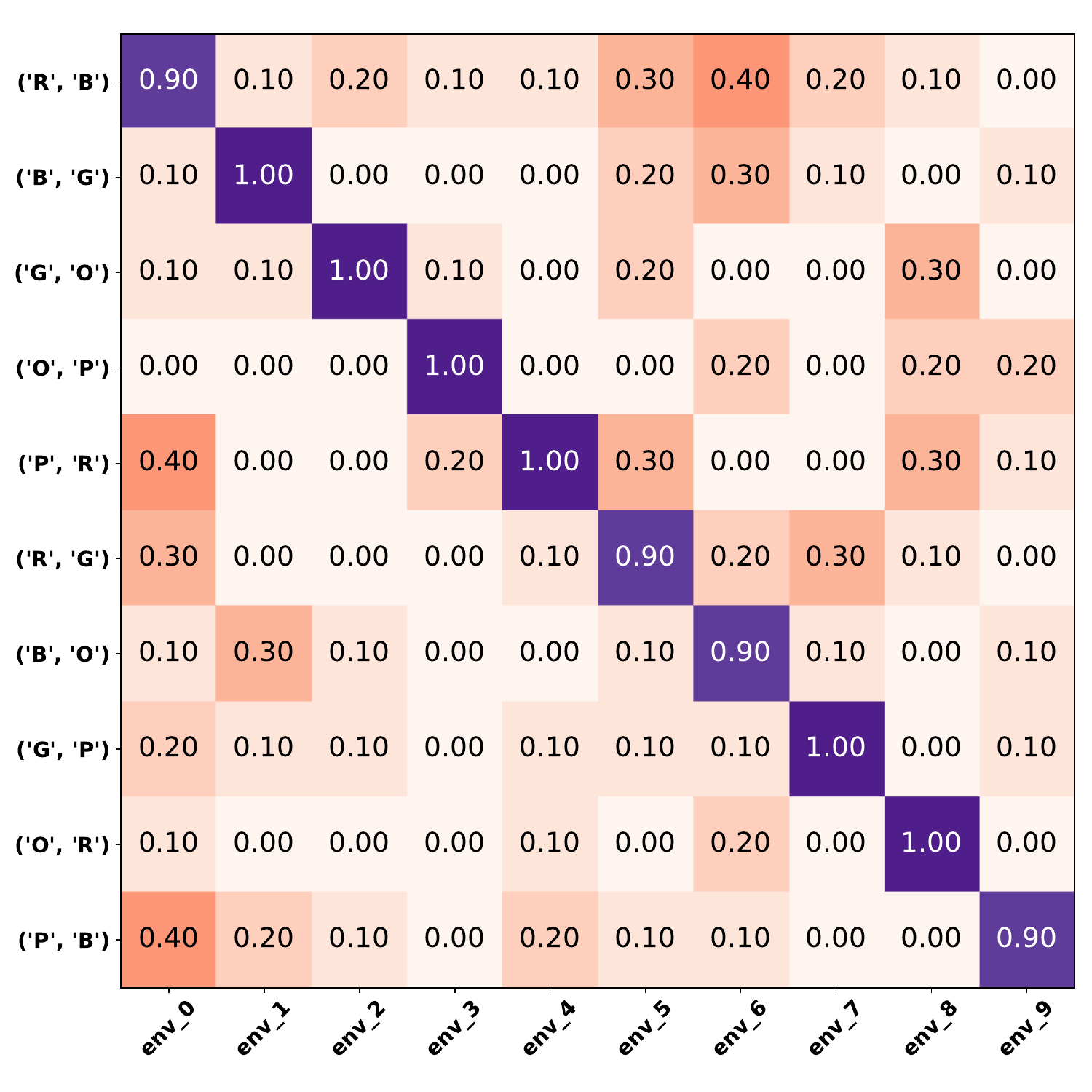} &
		\includegraphics[width=0.475\textwidth]{figures/gridworld/lambda_confusion/SynPo-28} \\
		
		\textbf{(c) 30 Train and 70 Test } & \textbf{(d) 40 Train and 60 Test } \\
		\includegraphics[width=0.475\textwidth]{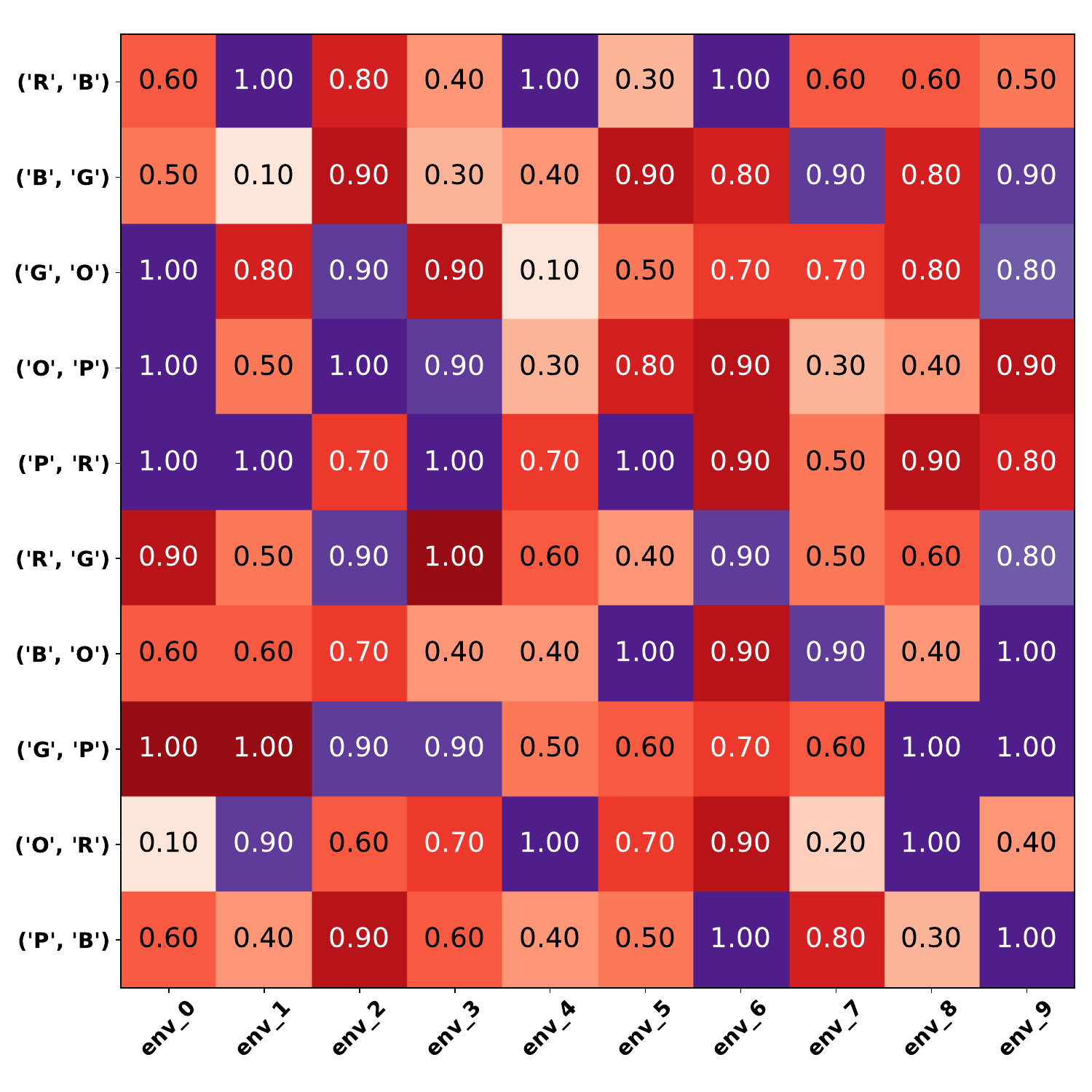} &
		\includegraphics[width=0.475\textwidth]{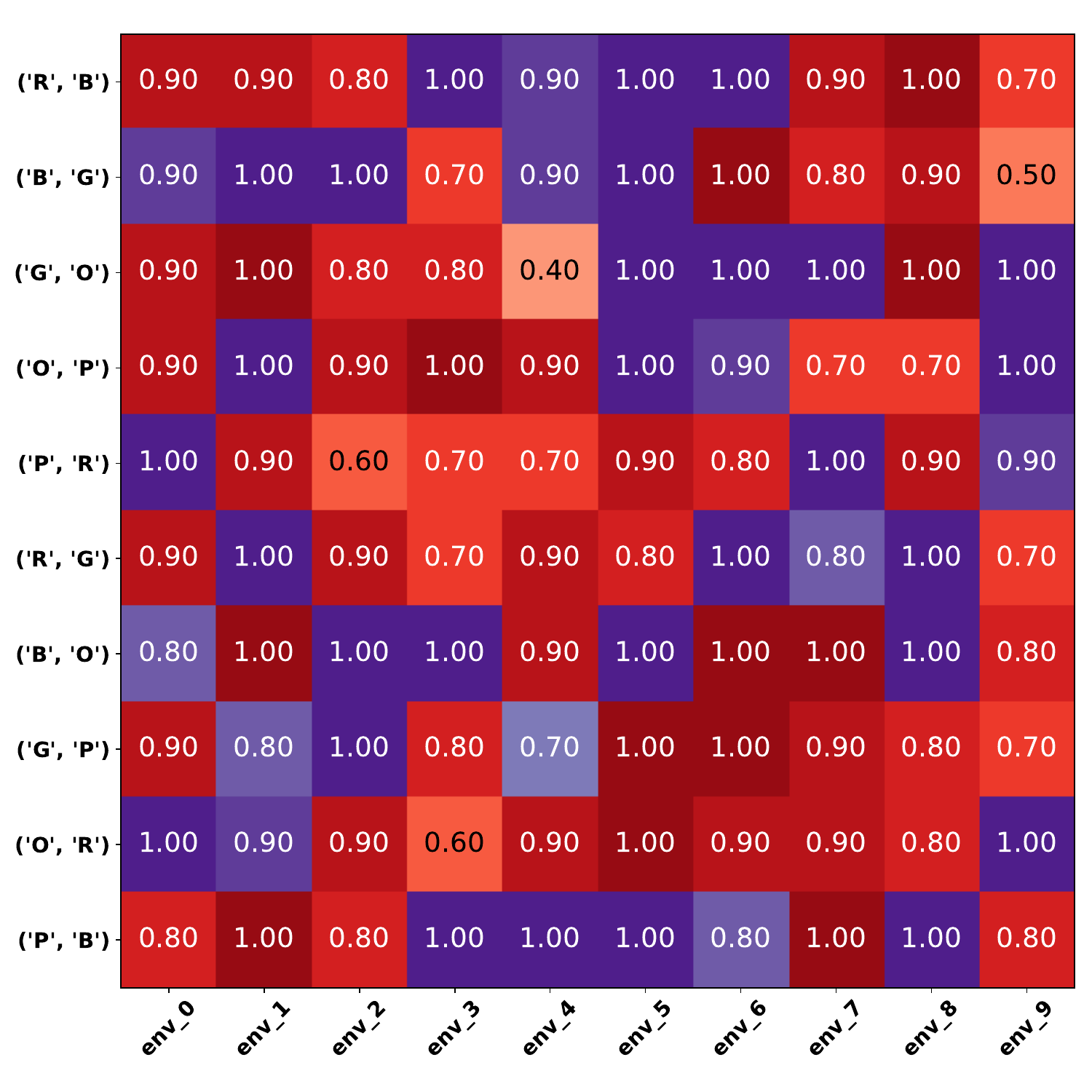} \\
	\end{tabular}
	\caption{Average test success rate on each environment-task combination. Blue grids represent seen combinations and red grids represent unseen combinations}
\end{figure}

\begin{figure}[h]
	\small
	\centering
	\label{tab:confusion2}
	\begin{tabular}{cc}
		\textbf{(a) 50 Train and 50 Test } & \textbf{(b) 60 Train and 40 Test } \\
		\includegraphics[width=0.475\textwidth]{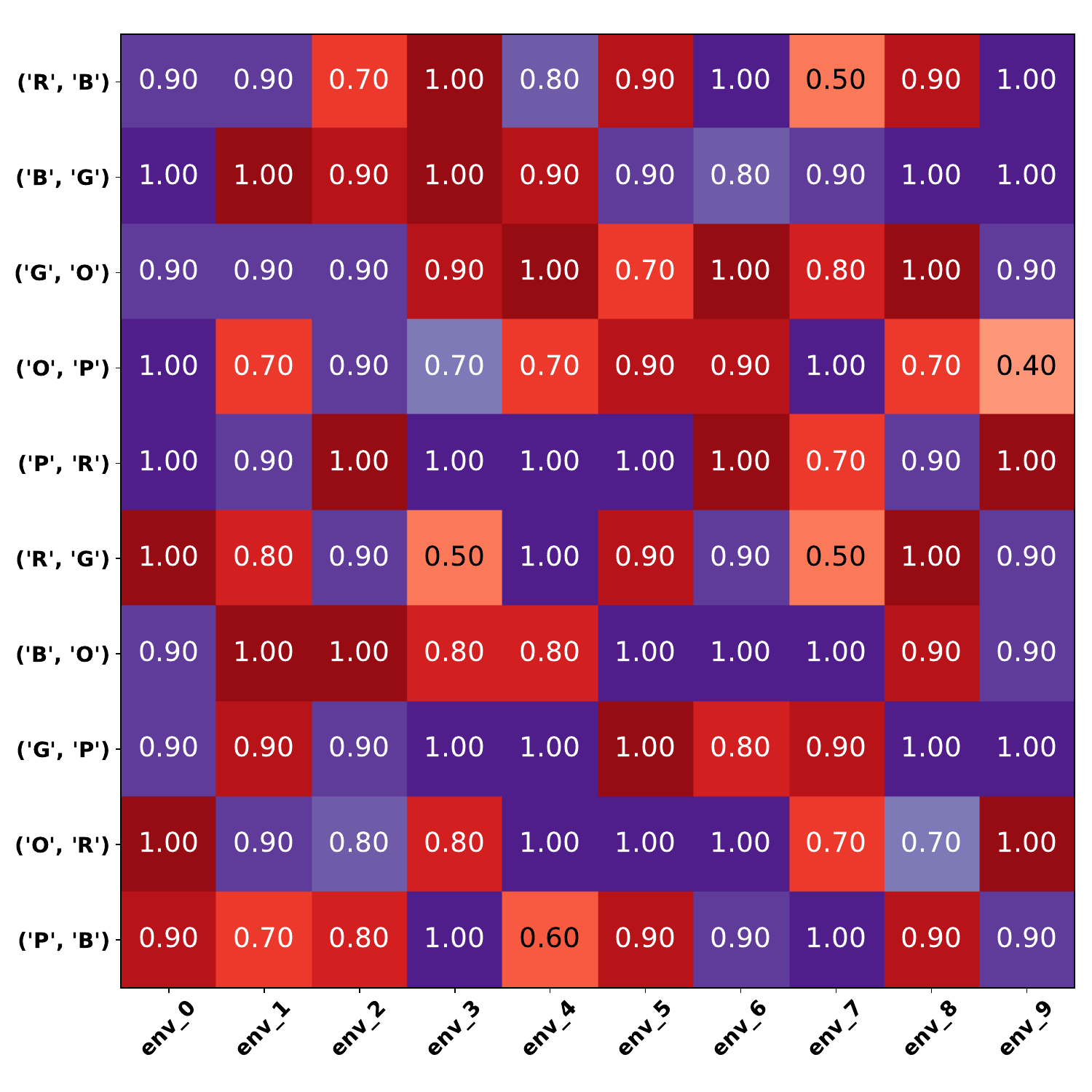} &
		\includegraphics[width=0.475\textwidth]{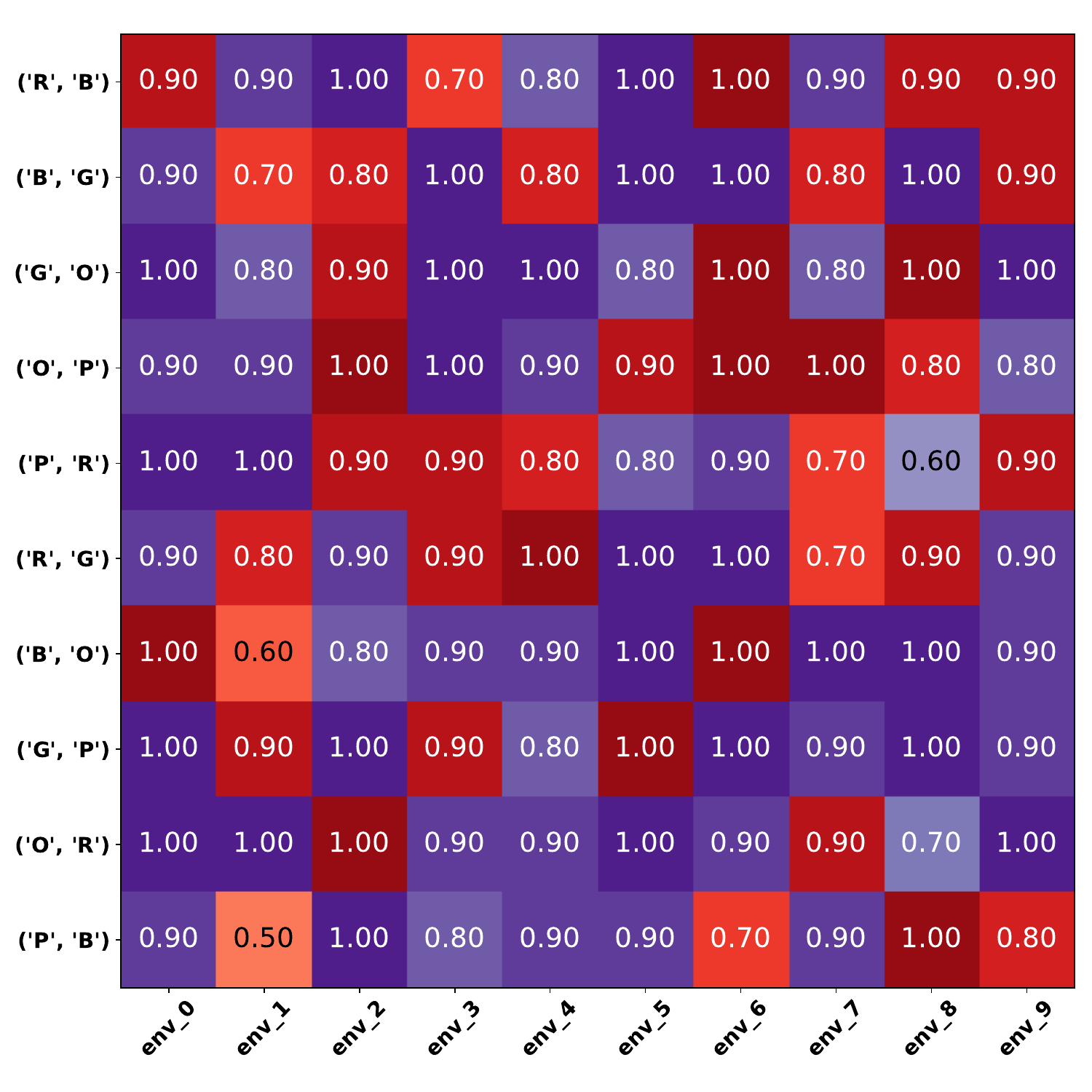} \\
		
		\textbf{(c) 70 Train and 30 Test } & \textbf{(d) 80 Train and 20 Test } \\
		\includegraphics[width=0.475\textwidth]{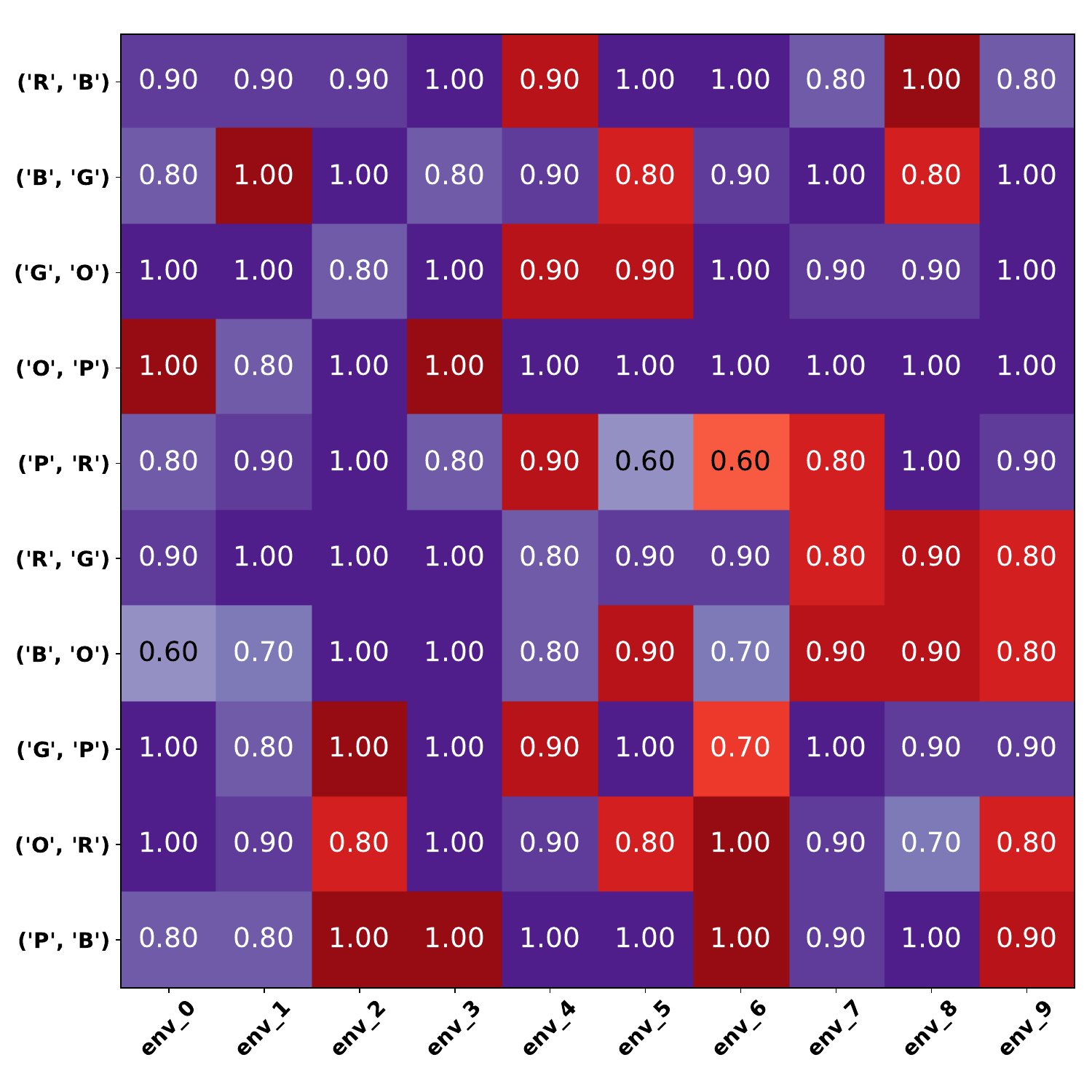} &
		\includegraphics[width=0.475\textwidth]{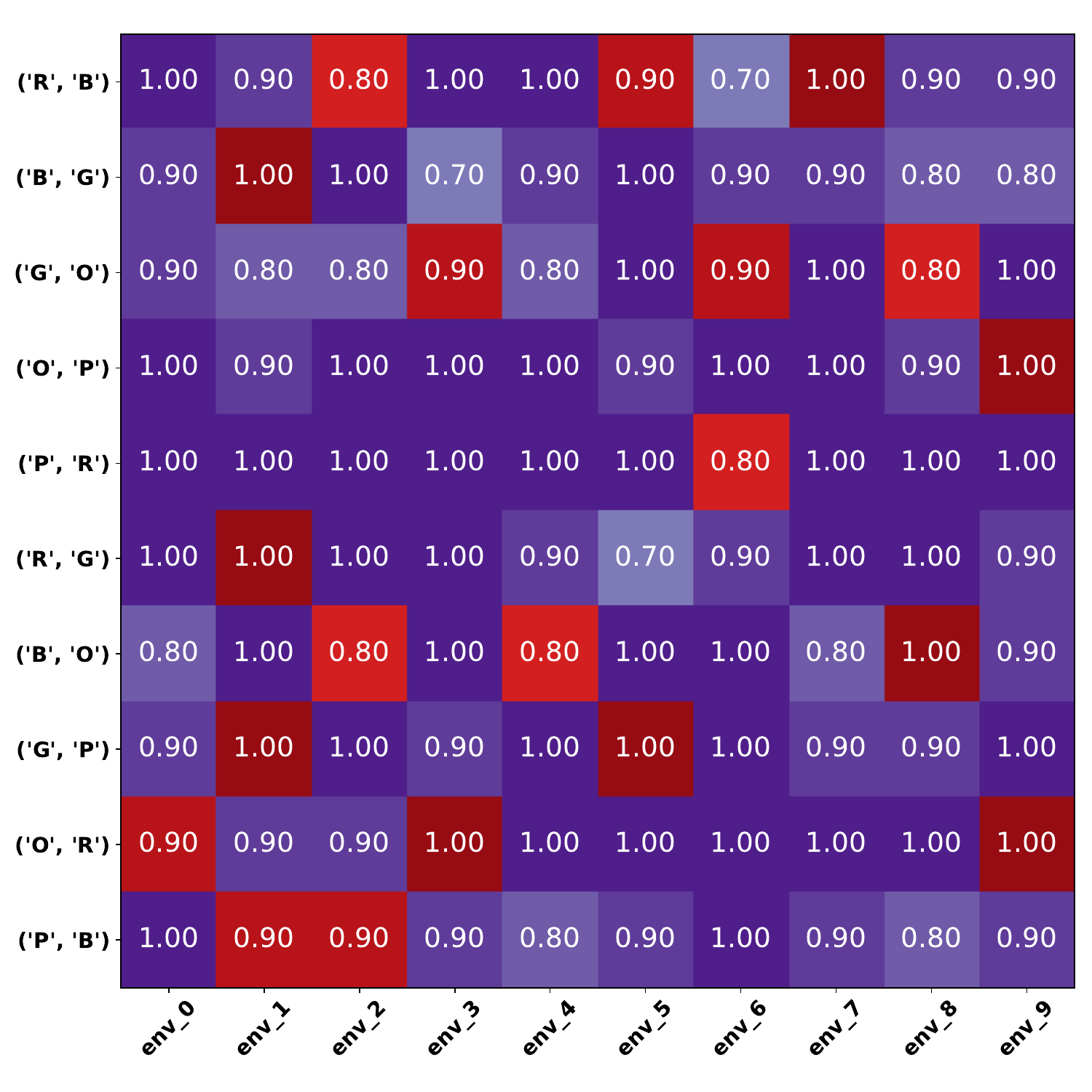} \\
	\end{tabular}
	\caption{Average test success rate on each environment-task combination. Blue grids represent seen combinations and red grids represent unseen combinations}
\end{figure}

As mentioned in the main text, here we include the complete splits for the transfer learning study (Experiments evaluated the transfer learning result w.r.t. ratio \# of seen vs.\# of total). The success rate of our method on each pair is marked on the matrices. The full success rate matrices are shown as Figure~4 and Figure~5.

\begin{figure}[ht]
	\centering
	\includegraphics[width=0.675\textwidth]{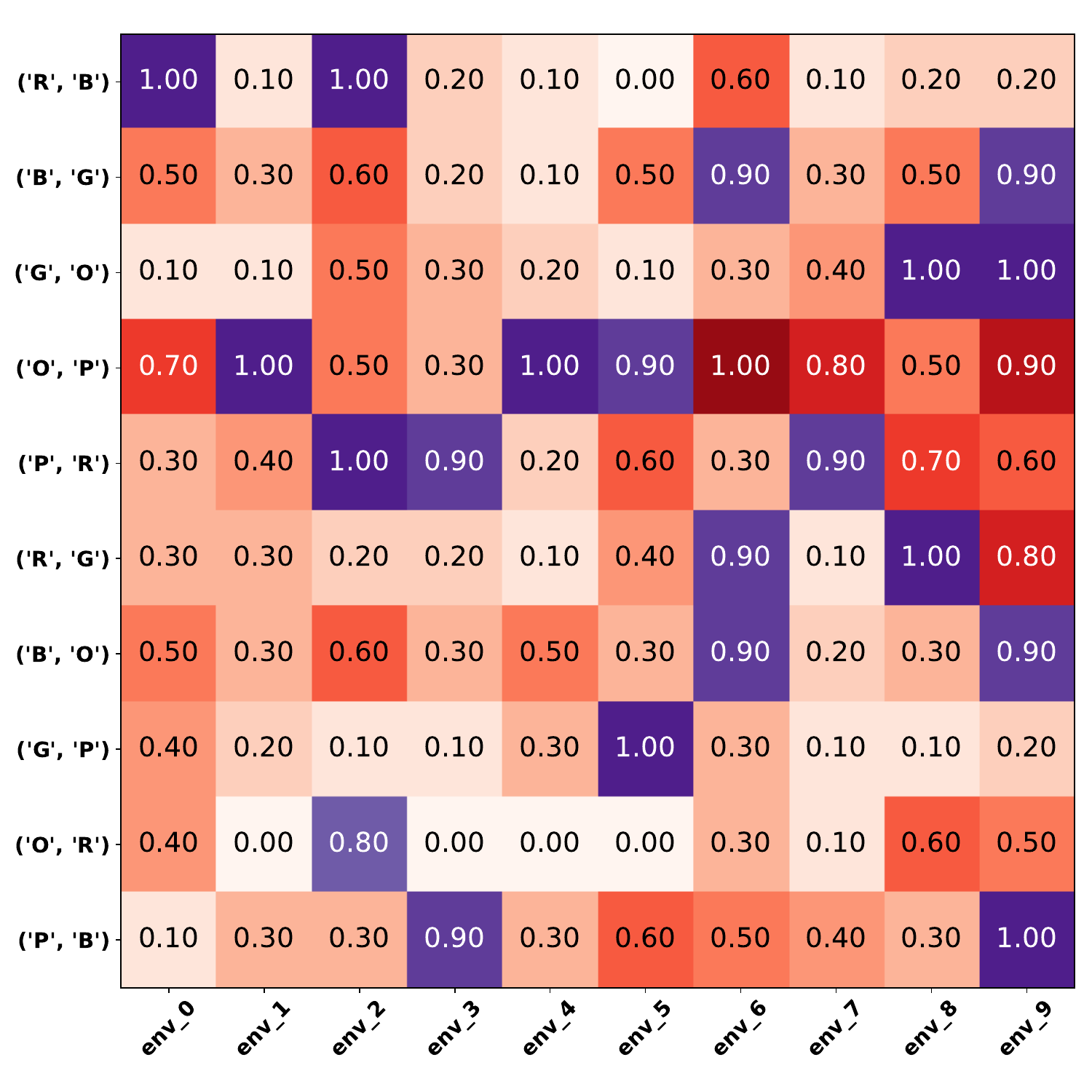}
	\caption{Case study for a situation when the ratio of \# of combinations seen and the total is 0.2}
	\label{fig:casestudy}
\end{figure}

Specifically we case study the situation when this ratio is $0.2$. The detailed transfer learning performance is shown as Figure~\ref{fig:casestudy}. Here each row corresponds to a task and each column corresponds to an environment. The red grids represents the unseen pairs and the purple grids represents the seen pairs. We mark the average success rate (over 100 runs of evaluations) in the grid to better quantitatively identify the performance at a pair of (\env, \task). The darker the color of a grid is, the better the corresponding performance. We can see that with the row ``(O, R)'' and column ``env\_0'', although only entry along the row and column is seen by the model, the transfer learning performance does not fail completely. Instead, many entries along the row and column have a superior success rate. This supports our claim about disentanglement of the environment and task embedding, and at the same time indicates the success in the learning compositionality. 

\subsection{Details on Experiments of transfer setting 2 and setting 3}

In this section, we describe the details of transfer learning settings. In both the setting 2 of ``Incremental learning of small pieces and integrating knowledge later'' and setting 3 of ``Learning in giant jumps and connecting dots'', we fix all parameters of the policy basis pre-trained on $P$ and fine-tune the network to learn new (randomly initialized) embeddings for environments and tasks. In this stage, we use only one demonstration from each (\env, \task) pair to fine-tune the embedding and find that our network is able to generalize to new environment or/and task. 

\begin{figure*}[ht]
	\centering
	\begin{tabular}{cc}
		\includegraphics[width=0.475\textwidth,trim={2.25cm 0 2cm 1.5cm},clip]{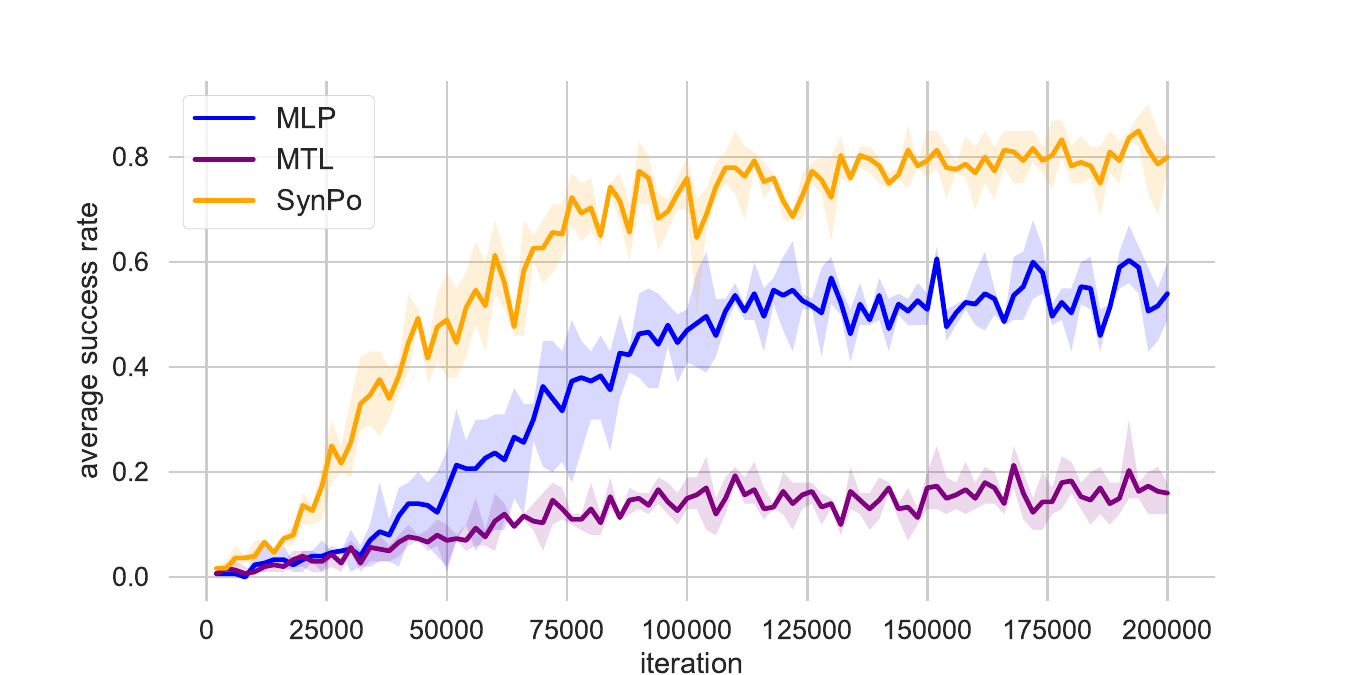} &
		\includegraphics[width=0.475\textwidth,trim={2.25cm 0 2cm 1.5cm},clip]{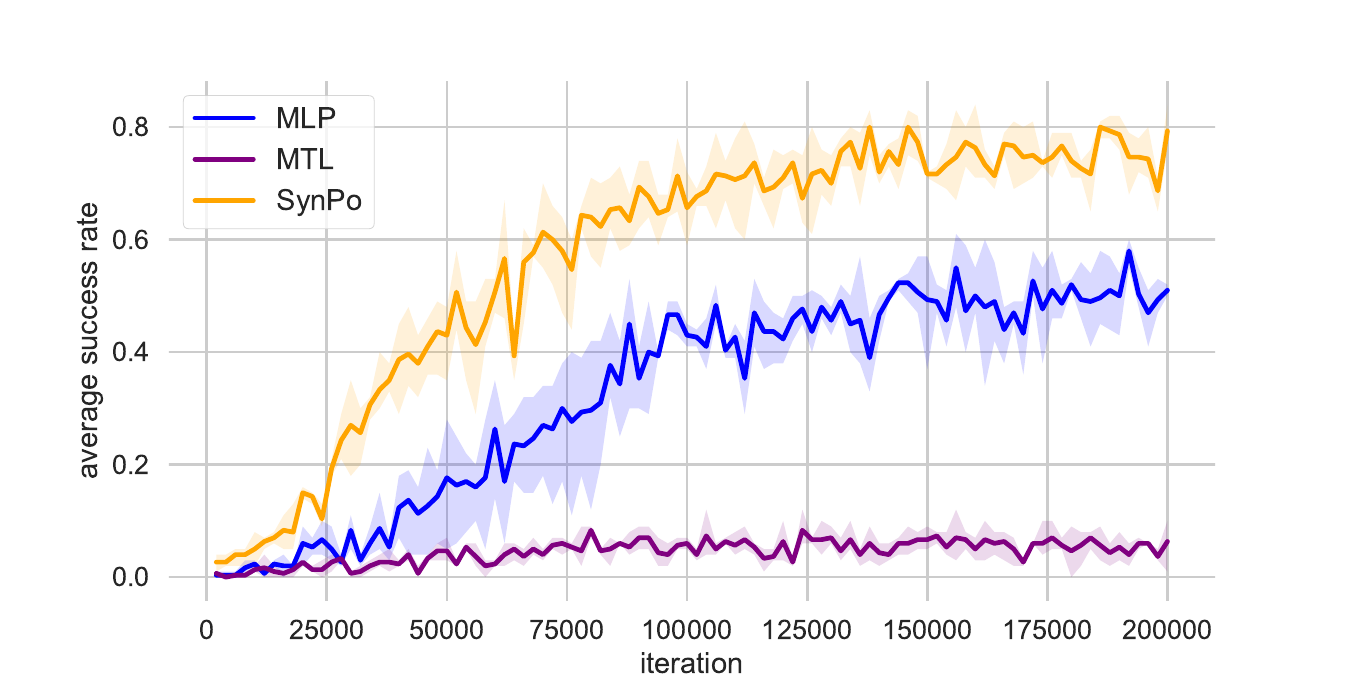} \\
		(a) AvgSR. on \textsc{Seen} Split & (b) AvgSR. on \textsc{Unseen} Split \\
		\includegraphics[width=0.475\textwidth,trim={2.25cm 0 2cm 1.5cm},clip]{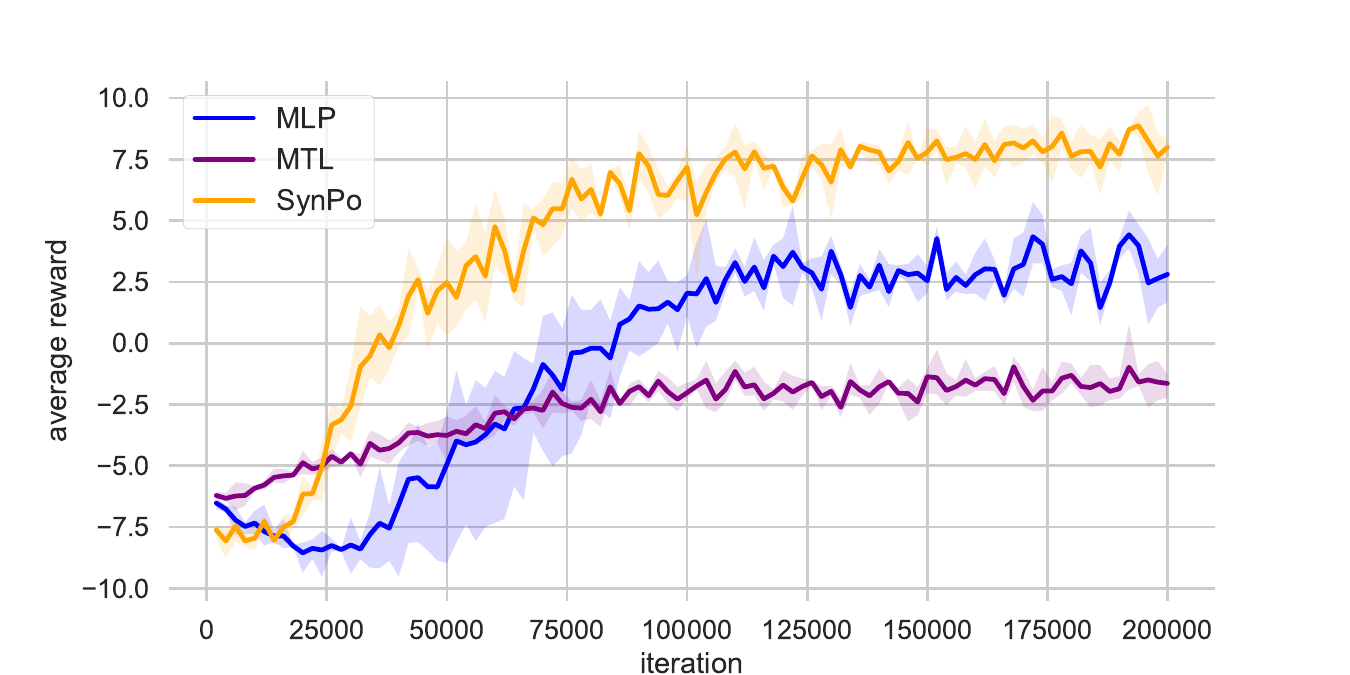} &
		\includegraphics[width=0.475\textwidth,trim={2.25cm 0 2cm 1.5cm},clip]{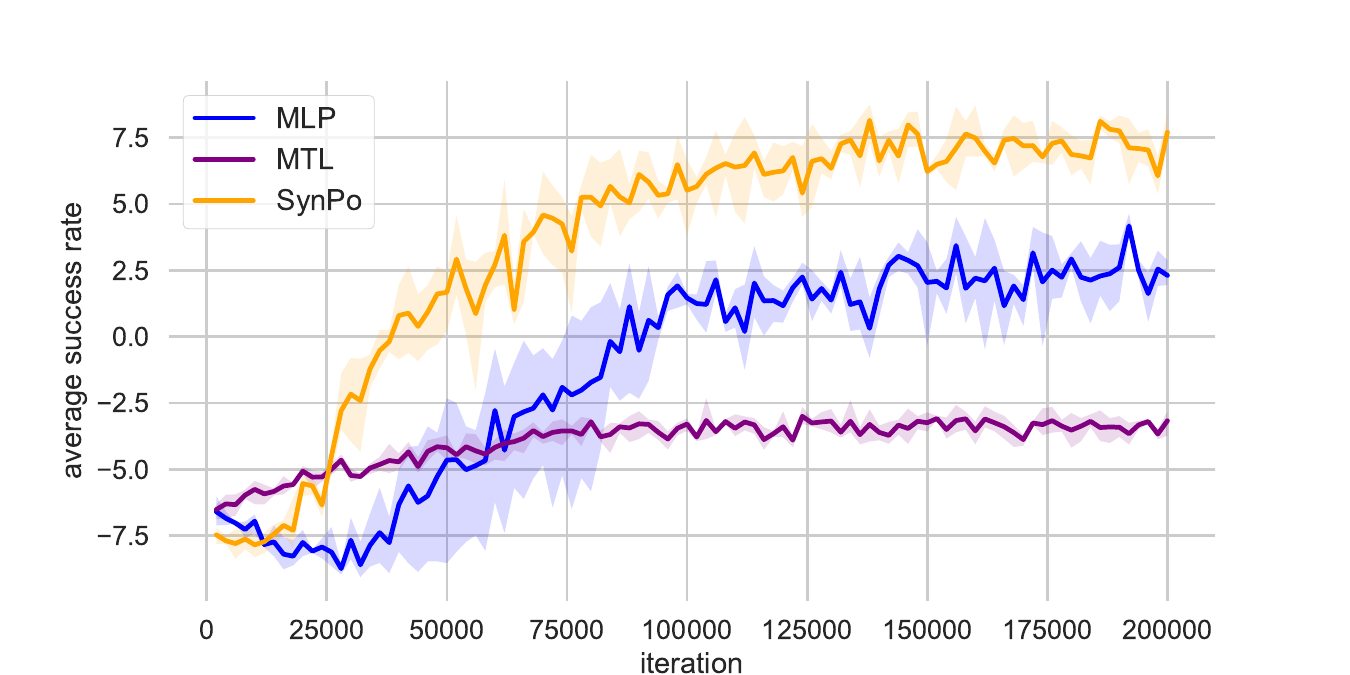} \\
		(c) AvgReward on \textsc{Seen} Split & (d) AvgReward on \textsc{Unseen} Split \\
	\end{tabular}
	\caption{\textbf{Results of ``A blind agent scenario'' on \gridworld with window size of 0. } (a)-(b): Comparison between average success rate (ASR.) of algorithms on {seen} split and {unseen} split. (c)-(d): Comparison between average accumulated reward (AvgReward.) of algorithms in each episode on {seen} split and {unseen} split. Results are reported on the setting with \textit{$|\calE| = 20$ and $|\calT| = 20$}. For each intermediate performance, we sample 100 (\env, \task) combinations and test one configuration to evaluate the performances. We evaluate models trained with 3 random seeds and report results in terms of the mean AvgSR and its standard deviation. }
	\label{fig:success_rate}
	\vskip-1em
\end{figure*}

Concretely, we randomly initialize the 10 new environment embeddings and the 10 new task embeddings for additional learning. In the transfer setting 2, we sample only one expert trajectory as demonstration data for each (\env, \task) pair in the upper right and lower left quadrant. In the transfer settings 3, we sample only one expert trajectory as demonstration data for each (\env, \task) pair in the lower right quadrant. Following the same routine Algorithm 1, we train the embeddings for 10000 iterations and then test the performance of models on the entire matrix of (\env, \task) pairs. The result is shown as Figure~\ref{fig:extensibility}. Besides what we have mentioned in the main text, we plot a more visually discernible success rate matrices as Figure~\ref{fig:extensibility} (a) and (b). We observe that in both cases, transfer learning across the task axis is easier comparing to the environment axis, given the results.  

\subsection{An extreme studies about the effectiveness of environment embeddings.}

\begin{figure}[h]
	\centering
	\begin{tabular}{c}
		\includegraphics[width=0.72\textwidth]{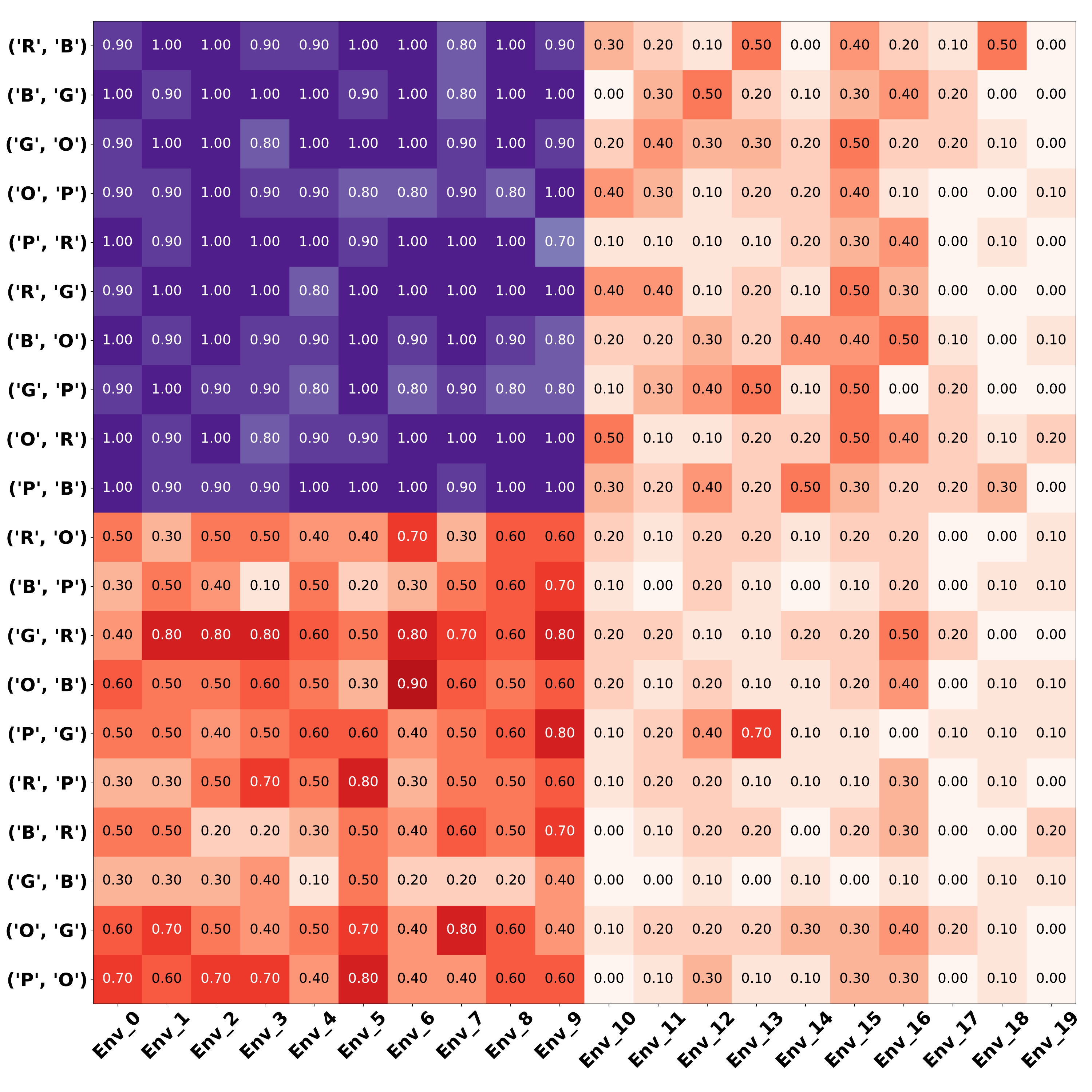} \\
		(a) Transfer Setting 2 \\
		\includegraphics[width=0.72\textwidth]{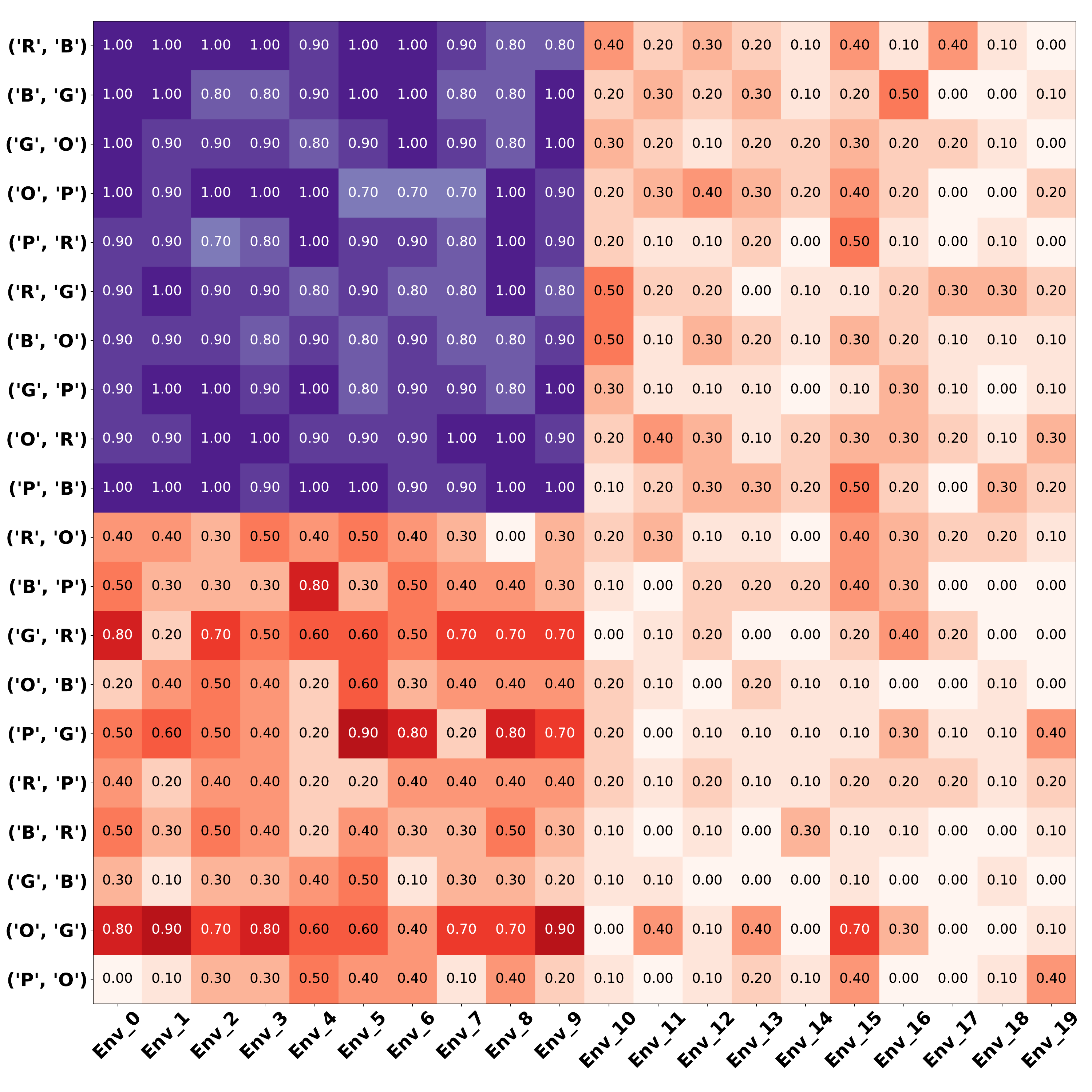} \\
		(b) Transfer Setting 3 \\
	\end{tabular}
	\caption{\small Visualizing the effectiveness transferring.  Average success rates are marked in the grid (more visually discernible plots are in the Suppl. Materials). The purple cells are from $Q$ set and red cells represents the rest. The darker the color is, the better the corresponding performance.}
	\label{fig:extensibility}
\end{figure}

As mentioned in the main text, to study the effectiveness of the environment embedding, we run an additional experiment as a sanity check. In this setting, we made agent's observation window size to be 1, which made agent only capable of seeing itself and the location of treasures on the map, without any knowledge about the maze. We denote this agent as a ``blind'' agent. Therefore, such a agent would need to remember the structure of the maze to perform well under this circumstance. We follow our original imitation training process as well as evaluation process and tested three representative methods in this setting, and plot the results as Table~\ref{tab:gridworld_blind}. As we have expected, we observe that algorithms such as MTL which do not distinguish between environments would fail severely. It could still success in some cases such as the treasures are generated at the same room as the agent, or very close by. With the additional environment embedding, a simple algorithm such as MLP could significantly outperforms this degenerated multi-task model. In addition, \ourmethod can achieve almost as good as it was in the normal circumstance, demonstrating its strong capability in memorizing the environment.

\begin{table}[h]
	\centering
	\setlength{\tabcolsep}{5pt}
	\caption{Performance of SynPo, MTL and MLP on \gridworld (\textsc{Seen}/\textsc{Unseen}=144/256) with window size = 0. All algorithms trained are trained using three random seeds and reported with mean and std. }
	\begin{tabular}{c | cc | c}
		Method & MLP & MTL & \ourmethod \\ \hlineB{3}
		AvgSR. (\textsc{Seen}) & 56.8 $\pm$ 0.9\% & 16.4 $\pm$ 0.4\% & \textbf{80.9 $\pm$ 1.5 \%} \\
		AvgSR. (\textsc{Unseen}) & 51.8 $\pm$ 1.7\% & 6.1 $\pm$ 0.2\% & \textbf{76.8 $\pm$ 1.4\%} \\
	\end{tabular}
	\label{tab:gridworld_blind}
\end{table}


\def\VizImageWidth{0.835in}
\def\VizImageWidth{0.835in}

\begin{figure}[h]
	\vspace{-0.15in}
	\small
	\centering
	\begin{tabular}{cccc}
		\includegraphics[width=0.245\textwidth]{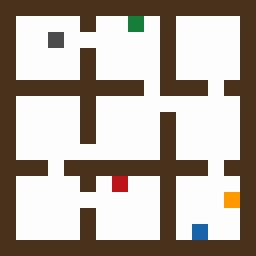} &
		\includegraphics[width=0.245\textwidth]{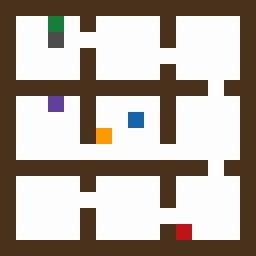} &
		\includegraphics[width=0.245\textwidth]{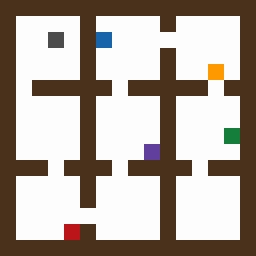} &
		\includegraphics[width=0.245\textwidth]{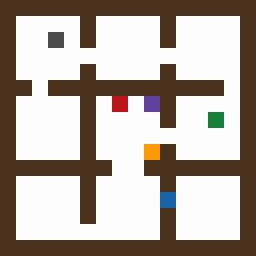} \\
		
		Env 1 & Env 2 & Env 3 & Env 4 \\
		
		\includegraphics[width=0.245\textwidth]{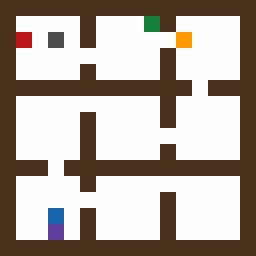} &
		\includegraphics[width=0.245\textwidth]{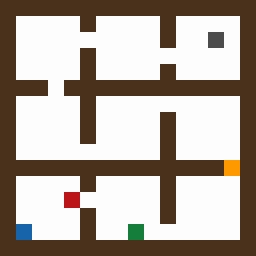} &
		\includegraphics[width=0.245\textwidth]{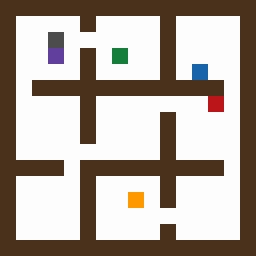} &
		\includegraphics[width=0.245\textwidth]{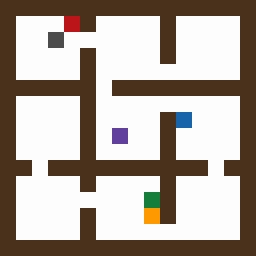} \\
		
		Env 5 & Env 6 & Env 7 & Env 8 \\
		
		\includegraphics[width=0.245\textwidth]{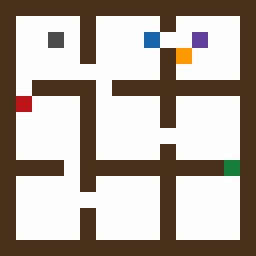} &
		\includegraphics[width=0.245\textwidth]{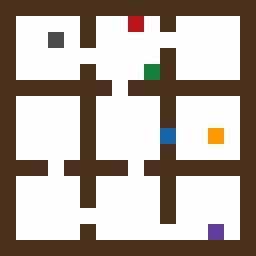} &
		\includegraphics[width=0.245\textwidth]{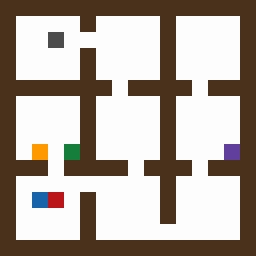} &
		\includegraphics[width=0.245\textwidth]{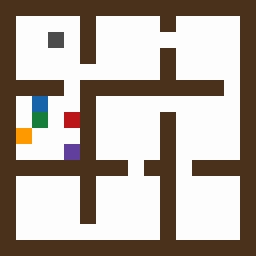} \\
		
		Env 9 & Env 10 & Env 11 & Env 12 \\
		
		\includegraphics[width=0.245\textwidth]{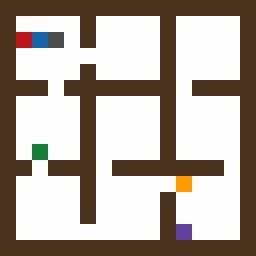} &
		\includegraphics[width=0.245\textwidth]{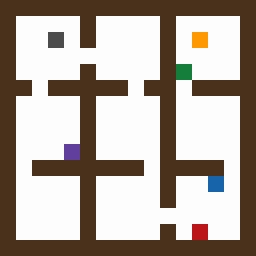} &
		\includegraphics[width=0.245\textwidth]{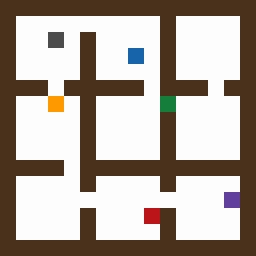} &
		\includegraphics[width=0.245\textwidth]{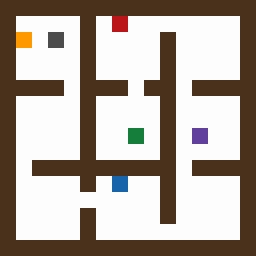} \\
		
		Env 13 & Env 14 & Env 15 & Env 16 \\
		
		\includegraphics[width=0.245\textwidth]{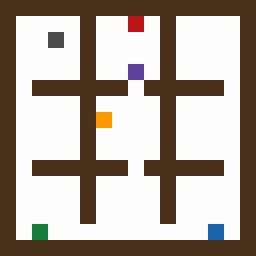} &
		\includegraphics[width=0.245\textwidth]{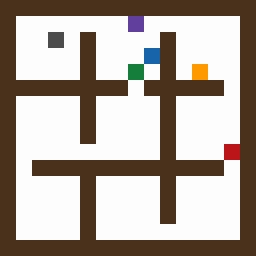} &
		\includegraphics[width=0.245\textwidth]{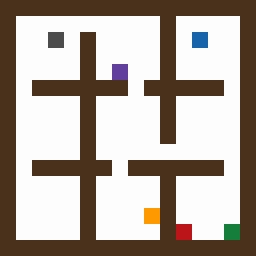} &
		\includegraphics[width=0.245\textwidth]{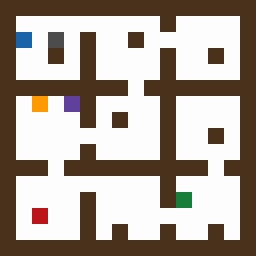} \\
		
		Env 17 & Env 18 & Env 19 & Env 20 \\
		
	\end{tabular}
	\caption{Visualization of the environments we used for \gridworld experiments. The environments we used are very different against each other, thus placed a substantial challenge for agent to generalize. (Note that agent's and objects' positions are randomized.)}
	\label{fig:gridworld_maps}
	\vspace{-0.05in}
\end{figure}

\end{document}